\documentclass[authoryear,preprint,12pt]{elsarticle}

\usepackage{graphicx}

\usepackage{amssymb,amsthm}
\usepackage{graphicx} 
\usepackage{epsfig} 
\usepackage{subcaption}
\usepackage{amsmath,amsfonts} 
\usepackage{bm}
\usepackage[margin=0.85in]{geometry}
\usepackage{lineno}

\usepackage{psfrag}

\usepackage{hyperref}
\hypersetup{
     colorlinks   = true,
     citecolor    = gray
}

\renewcommand*{\contentsline}[3]{\csname l@#1\endcsname{#2}{}}
\usepackage{tocloft}

\usepackage{color}

\usepackage{enumitem}

\usepackage{epsfig} 
\usepackage{amsmath} 
\usepackage{amssymb}  
\usepackage{microtype}
\usepackage{epstopdf}
\usepackage{booktabs}
\usepackage{breqn}
\usepackage{bm}

\usepackage{multirow}
\usepackage[ruled,vlined,linesnumbered]{algorithm2e}

\usepackage{tikz}
\usetikzlibrary{shapes,arrows,chains}
\tikzset{middlearrow/.style={
        decoration={markings,
            mark= at position 0.6 with {\arrow{#1}} ,
        },
        postaction={decorate}
    }
}

\usepackage{multirow}
\usepackage{url}
\usepackage{makecell}
\usepackage{relsize}
\usepackage{adjustbox}

\usepackage{mathtools}
\DeclareMathAlphabet{\pazocal}{OMS}{zplm}{m}{n}
\makeatletter
\newcommand{\vast}{\bBigg@{3}}
\makeatother
\makeatletter
\renewcommand*\env@matrix[1][\arraystretch]{%
  \edef\arraystretch{#1}%
  \hskip -\arraycolsep
  \let\@ifnextchar\new@ifnextchar
  \array{*\c@MaxMatrixCols c}}
\makeatother

\newcolumntype{L}[1]{>{\raggedright\let\newline\\\arraybackslash\hspace{0pt}}m{#1}}
\newcolumntype{C}[1]{>{\centering\let\newline\\\arraybackslash\hspace{0pt}}m{#1}}
\newcolumntype{R}[1]{>{\raggedleft\let\newline\\\arraybackslash\hspace{0pt}}m{#1}}

\begin{document}

\begin{frontmatter}

\title{Critical ride comfort detection for automated vehicles}

\author[ETHZ]{Alexander Genser\corref{cor1}}
\ead{gensera@ethz.ch}

\author[AIT]{Roland Spielhofer}
\ead{roland.spielhofer@ait.ac.at}

\author[AVL]{Philippe Nitsche}
\ead{philippe.nitsche@avl.com}

\author[ETHZ]{Anastasios Kouvelas}
\ead{akouvela@ethz.ch}

\cortext[cor1]{Corresponding author. Tel.:  +41-44-632-75-19.}

\address[ETHZ]{Institute for Transport Planning and Systems, Department of Civil, Environmental and\\Geomatic Engineering, ETH Zurich, CH-8093 Zurich, Switzerland}

\address[AIT]{AIT -- Austrian Institute of
Technology GmbH, Center for Low-Emission Transport, AT-1210 Vienna, Austria}

\address[AVL]{AVL List GmbH, Engineering and Technology Powertrain Systems, AT-8020 Graz, Austria}


\begin{abstract}
 In a future connected vehicle environment, an optimized route and motion planning should not only fulfill efficiency and safety constraints but also minimize vehicle motions and oscillations, causing poor ride comfort perceived by passengers. This work provides a framework for a large-scale and cost-efficient evaluation to address AV's ride comfort and allow the comparison of different comfort assessment strategies. The proposed tool also gives insights to comfort data, allowing for the development of novel algorithms, guidelines, or motion planning systems incorporating passenger comfort. A vehicle-road simulation framework utilizable to assess the most common ride comfort determination strategies based on vehicle dynamics data is presented. The developed methodology encompasses a road surface model, a non-linear vehicle model optimization, and Monte Carlo simulations to allow for an accurate and cost-efficient generation of virtual chassis acceleration data. Ride comfort is determined by applying a commonly used threshold method and an analysis based on ISO 2631. The two methods are compared against comfort classifications based on empirical measurements of the International Roughness Index (IRI). A case study with three road sites in Austria demonstrates the framework's practical application with real data and achieves high-resolution ride comfort classifications. The results highlight that ISO 2631 comfort estimates are most similar to IRI classifications and that the thresholding procedure detects preventable situations but also over- or underestimates ride comfort. Hence, the work shows the potential risk of negative ride comfort of AVs using simple threshold values and stresses the importance of a robust comfort evaluation method for enhancing AVs' path and motion planning with maximal ride comfort. 
\end{abstract}


\end{frontmatter}



\section{INTRODUCTION}
\label{sec:1-Intro}
The deployment of Automated vehicles (AV) involves several challenges, such as the change of the human driver's role. With AVs, the role of a human driver shifts towards the one of a vehicle passenger. As a result, the importance of user acceptance of such technologies arises and should not be neglected during the design procedure. Although there is a broad spectrum of ongoing research (e.g.,\ improvements and development of Advanced Driver Assistance Systems (ADAS)), not much attention has been devoted to the area of ride comfort. In this paper, ride comfort is defined as the overall perceived comfort by the passengers caused by vibrations in the passenger compartment. While some oscillations can be minimized by vehicle design, usually part of a NVH (noise, vibration and harshness) optimization, some are actuated by a rough road surface. Recent studies have pointed out that passengers in an AV are more likely to experience deficient ride comfort than passengers in a maneuvering non-automated vehicle~\cite{ref:rosenzweig, ref:smyth, ref:sawabe}. Hence, the data pool an AV utilizes for efficient and user-friendly motion planning should be enhanced with (a) comfort data representing current critical sections in a road network and (b) a suitable and robust method that does not allow over- or underestimation of critical comfort. 
To the best of the authors' knowledge, this work is the first to study AVs' ride comfort with an extensive data pool. Besides, studies tackled the correlation of different methods (e.g., ISO 2631 and IRI) but did not consider the specific comfort results and their importance for passenger comfort in AVs.

In order to tackle the challenge of mitigating deficient ride comfort, this paper compares different methodologies for determining ride comfort. First, we propose a sub-microscopic simulation framework allowing for a cost-efficient and generic generation of accurate vehicle dynamics from AVs. A physical vehicle model with accurate vehicle dynamics implementation and an Adaptive Cruise Control (ACC) has been applied. To calibrate the non-linear vehicle model for an optimum correlation to real behaviour, we utilize real-world vehicle dynamics measurements and solve an optimization problem to minimize the modeling error. After calibration and determination of model parameters, Monte Carlo (MC) simulations with Latin Hypercube Sampling (LHS) are performed. A representative set of input samples for the vehicle's speed profile and the lateral position in the lane is obtained to model different driver behaviors. Also, we sample the friction coefficient of the road surface as a proxy for different weather conditions. Finally, a representative number of simulations allows the utilization of output data to evaluate the ride comfort based on (a) most recent definitions of threshold values, (b) the guidelines from the International Standardization Organization (ISO) ISO 2631, and (c) the International Roughness Index (IRI). For proof of concept, the methodology is applied to three test sections in Austria. 

The remainder of this paper is organized as follows: Fundamental works on ride comfort determination from vehicle dynamics and the research gap towards AVs are presented in Section~\ref{sec:back_motivation}. Sections~\ref{sec:sim_framework} and~\ref{sec:model_val_opti} explain the architecture of the simulation framework and the model optimization procedure, respectively. The strategies to determine ride comfort estimates is introduced in Section~\ref{sec:ride_comfort}. Finally, the conducted case study and the obtained results are presented in Section~\ref{sec:case_study}. The paper closes with a conclusion and potential future research in Section~\ref{sec:diss_concl}. 

\section{BACKGROUND AND CONTRIBUTIONS}
\label{sec:back_motivation}
Besides all novel technical methodologies and improvements in the automotive domain (e.g., ADAS), less attention is paid to human comfort that is potentially influenced by the introduction of AVs on the market~\cite{ref:kyriakidis}. The number of publications within the last decade in the AV domain shows a substantial increase. However, at the same time, the study of the overall comfort of humans during a vehicle trip in an AV is a relatively new research field~\cite{ref:rosenzweig, ref:du}. When vibrations in a vehicle exceed certain thresholds, the occupants feel discomfort, and ride quality decreases. Nevertheless, when these vibrations are low, the human body feels relatively comfortable, and negative influences on well-being, health, or motion sickness are insignificant.

In a non-automated vehicle that is maneuvered by a human driver, the occupant can adjust the driving behavior upon the level of (dis)comfort. However, this is not possible in an AV when a-priori knowledge about the road infrastructure and traffic conditions are not available and/or sensing systems face difficulties (e.g., weather conditions, covered obstacles, or reflections). Hence, methodologies are needed to benchmark ride comfort objectively.~\cite{ref:Holzinger} have shown how to objectively assess vehicle driveability with the software AVL-DRIVE\texttrademark, focusing on ADAS/AD safety and comfort quality. However, there is limited literature about vibration-specific ride comfort evaluation caused by road roughness. Currently, most publications in this field address the public transportation sector. Therefore, in the following, this work analyses (a) studies of ride comfort determination that have been applied in the public transportation domain and/or (b) are relevant or applied to (automated) vehicles. 

Recent studies aimed at evaluating and improving passenger comfort utilize methods of different levels of sophistication. First, several publications derive a model (linear regressions are commonly applied) to predict the ride comfort by combining empirical kinematic measurements and subjective survey data. An extensive study from~\cite{ref:foerstberg} exploits the human response to different kinematic motions in trains. This work is one of the first, which combines empirical measurements (objective data) with data collected from a survey (subjective data). Regression analysis is performed, which finds that the most critical variables are motion dose from the lateral and roll acceleration in the horizontal plane. Due to the high low-frequency content of vehicle dynamic signals,~\cite{ref:foerstberg} proposes a model of provocation of motion sickness showing that motion sickness has a time decay or leakage. Along with the mentioned study, other works such as~\cite{ref:ZHANG, ref:eboli, ref:NGUYEN} combine objective and subjective data to determine ride comfort in public transportation vehicles. 
~\cite{ref:ZHANG} proposes a bus comfort model based on multiple linear regressions. Subjective parameters derived from a questionnaire and kinematic parameters (such as noise, thermal comfort, vibration, and acceleration) are considered. Also,~\cite{ref:NGUYEN} fits a linear regression model to explain the correlation between subjective passenger ratings and lateral acceleration and duration of a turning movement. However, the study also incorporates passenger posture. The presented performance of the model can be deemed modest.~\cite{ref:eboli} again merges subjective and objective measurement data and derives thresholds of comfort levels.

Secondly, threshold values are widely used for ride comfort determination, as the implementation is straightforward, and adjustments are inexpensive. In~\cite{ref:eriksson} different thresholds for the acceleration and jerk in public transport are collected, ranging from 1.00m/s$^2$ up to 1.47m/s$^2$ and  0.50m/s$^3$ up to 0.90m/s$^3$, respectively.~\cite{ref:elbanhawi} defines such thresholds ranging from 1.08m/s$^2$ and 1.47m/s$^2$ for the acceleration and 2.90m/s$^3$ as an upper bound for the jerk. The deviation of the jerk threshold can be explained by the fact that the study in~\cite{ref:eriksson} considers standing passengers in public transport, whereas~\cite{ref:elbanhawi} takes only sitting passengers into account.~\cite{ref:bae_threshold_comfort} provide the most recent extensive overview of threshold values with different categories of driving styles. The study is also one of the few compiling public transport and different driving styles for a personal vehicle. Upper and lower thresholds (i.e., acceleration and deceleration, respectively) for comfortable public transportation (PT), normal driving (ND), and aggressive driving (AD) are collected. The lateral acceleration bands for PT, ND and AD are defined by -0.90m/s$^2$ -- 0.90m/s$^2$, -4.00m/s$^2$ -- 4.00m/s$^2$, and -5.60m/s$^2$ -- 5.60m/s$^2$. For the longitudinal direction, the corresponding acceleration bands are listed with -0.90m/s$^2$ -- 0.90m/s$^2$, -2.00m/s$^2$ -- 1.47m/s$^2$, and -5.80m/s$^2$ -- 3.07m/s$^2$, for PT, ND and AD, respectively. In all the publications mentioned above, no threshold values for accelerations in the vertical direction are defined. 

Although thresholding might create comfortable vehicle trips on average, it does not guarantee the mitigation of discomfort in specific situations. Hence, thirdly, a more advanced approach is the analysis of the Power Spectral Density (PSD) of a given signal. By transforming the acceleration signals to the frequency domain with standard transformation methods (e.g.,\ Continuous Wavelet Transform (CWT)~\cite{ref:lee}) relevant frequency components can be analyzed. The ISO provides the guideline documents ISO 2631 that utilize the PSD approach to evaluate human exposure to whole-body vibrations. The guidelines provide a methodology to utilize acceleration signals in order to explore the effect of vibrations on (a) human health and comfort, (b) the probability of vibration perception, and (c) the emergence of motion sickness~\citep{ref:ISO2631}. ISO 2631 is widely applied in many different fields to evaluate human comfort/discomfort; e.g.,\ in structural engineering~\cite{ref:kralik2017}, industrial engineering~\cite{ref:zhang2018}, or automotive engineering~\cite{ref:xiao2020}. Specifically targeting ride comfort,~\cite{ref:CASTELLANOS} developed an embedded system to localize discomfort in public transportation networks and consequently derive and apply such thresholds. A comfort index measurement (the ISO 2631) and a jerk threshold detection are utilized to identify the comfort results; finally, the results are based on a threshold detection algorithm. Also,~\cite{ref:foerstberg2} investigates the International Union of Railways (UIC) standard to extend the evaluation to transition and circular curves. The UIC comfort tests are based on the ISO 2631 and specify the procedure for trains. The study aims again to find a correlation between passenger votes and UIC comfort parameters.    

Furthermore, another infrastructure-based approach to estimate ride comfort is the utilization of the International Roughness Index (IRI). The IRI utilizes a longitudinal road profile of a test section. The road roughness is measured by taking the fraction of a quarter-car model's output and the section length~\cite{ref:sayers_IRI_1995}. Consequently, an aggregated index for the sections' road roughness is given, which can also be utilized to determine ride comfort.~\cite{ref:nitsche_2014} evaluate pavement roughness by utilizing vehicle response measurements and machine learning models. Results show promising results at constant vehicle speed. One of the most recent insights into the interplay of ride comfort and IRI is given by~\cite{ref:mucka_iri_2020}. This work proposes IRI thresholds as a function of vehicle speed and road category. The methodology is based on finding a relationship between ISO 2631 and IRI. Results are compiled for nine different vehicles out of different categories (from a small family car, SUV, up to a cargo van).

As mentioned, vehicle ride comfort is only partially considered in the literature, and most of the studies focus on public transportation. Even less research has been conducted when looking at AVs.~\cite{ref:elbanhawi} provide a review article that reviews the state-of-the-art comfort assessment methods and also highlights their importance in path-planning algorithms for AVs. Further, the work emphasizes a potentially growing research gap of AV's comfort. To derive optimal path-planning, e.g.,~\cite{ref:sun2018} proposes a Pareto optimization problem that guarantees an optimal strategy by considering comfort (methodology based on the ISO 2631) and driving requirements. Additionally, AVs can offer different seating positions, which potentially influence the user's ride comfort. Therefore,~\cite{ref:salter} performed field experiments where test persons traveled in forward and rearward facing seats while performing office tasks. The results show a significant increase in the mean level of motion sickness with rearward sitting (especially in urban driving conditions). Hence, a conclusion can be drawn that there is a need for tools supporting large-scale ride comfort testing in AVs. 

The current work contains three novelties that extend the currently applied methodologies for ride comfort determination of AVs. First, a generic simulation framework is provided based on the work from~\cite{ref:nitsche_2018} and~\cite{ref:genser2019}, allowing for the generation of vehicle dynamics data with arbitrary resolutions based on a user-specified simulation scenario. Monte Carlo (MC) simulations with LHS sampling are applied. This methodology allows for a representative determination of vehicle dynamics by varying the speed profiles and the lateral vehicle position to model varying driving behaviors. To also include weather conditions in our simulations, we utilize the road surface's friction coefficient as a proxy. Hence, the framework allows for an in-detail evaluation  of available guidelines documents. 

Secondly, this is the first study that utilizes a high precision road surface model to determine (a) highly accurate vehicle dynamics from a simulation environment and (b) accurate ride comfort estimates. The data set for the surface model is measured with laser scanners, providing point data with a vertical accuracy of $\pm$0.1 mm. Moreover, a high precision positioning is ensured with post processing of the recorded trajectories using post processing of base station data . The surface model is incorporated into our framework using the standard OpenCRG specification. Consequently, we utilize the simulation set-up for a model optimization with measurement data from test drives on predefined road segments. The well-known Levenberg-Marquardt algorithm is utilized to optimize the non-linear system and minimize the simulator's model error~\cite{ref:Levenberg}. This procedure can be easily extended to different test vehicles, new test sites, and/or different driving patterns. 

Finally, we utilize high-resolution vehicle dynamics and extend the data to apply the most recent threshold values and the ISO 2631 procedure. Furthermore, the two methodologies are compared to comfort estimates based on the International Roughness Index (IRI) to underline the necessity of a robust and sophisticated determination procedure and draw the attention towards the importance of ride comfort data for motion planning of AVs to mitigate critical vehicle maneuvers. Comparison results are presented in the last section of the paper and derived insights are discussed.

\section{ARCHITECTURE OF SIMULATION FRAMEWORK}
\label{sec:sim_framework}
This section introduces the proposed simulation framework to derive accurate ride comfort data. First, the simulation framework's architecture and the corresponding components are introduced, followed by an in-detail description of the road surface modeling and the MC/LHS techniques. Note that the methodologies for the derivation of ride comfort data are part of the simulation framework, but a detailed overview is provided in Section~\ref{sec:ride_comfort}.

\subsection{Simulation environment}
\label{sec:sim_env}
\begin{figure}[!t]
    \centering
    \includegraphics[width=0.8\textwidth]{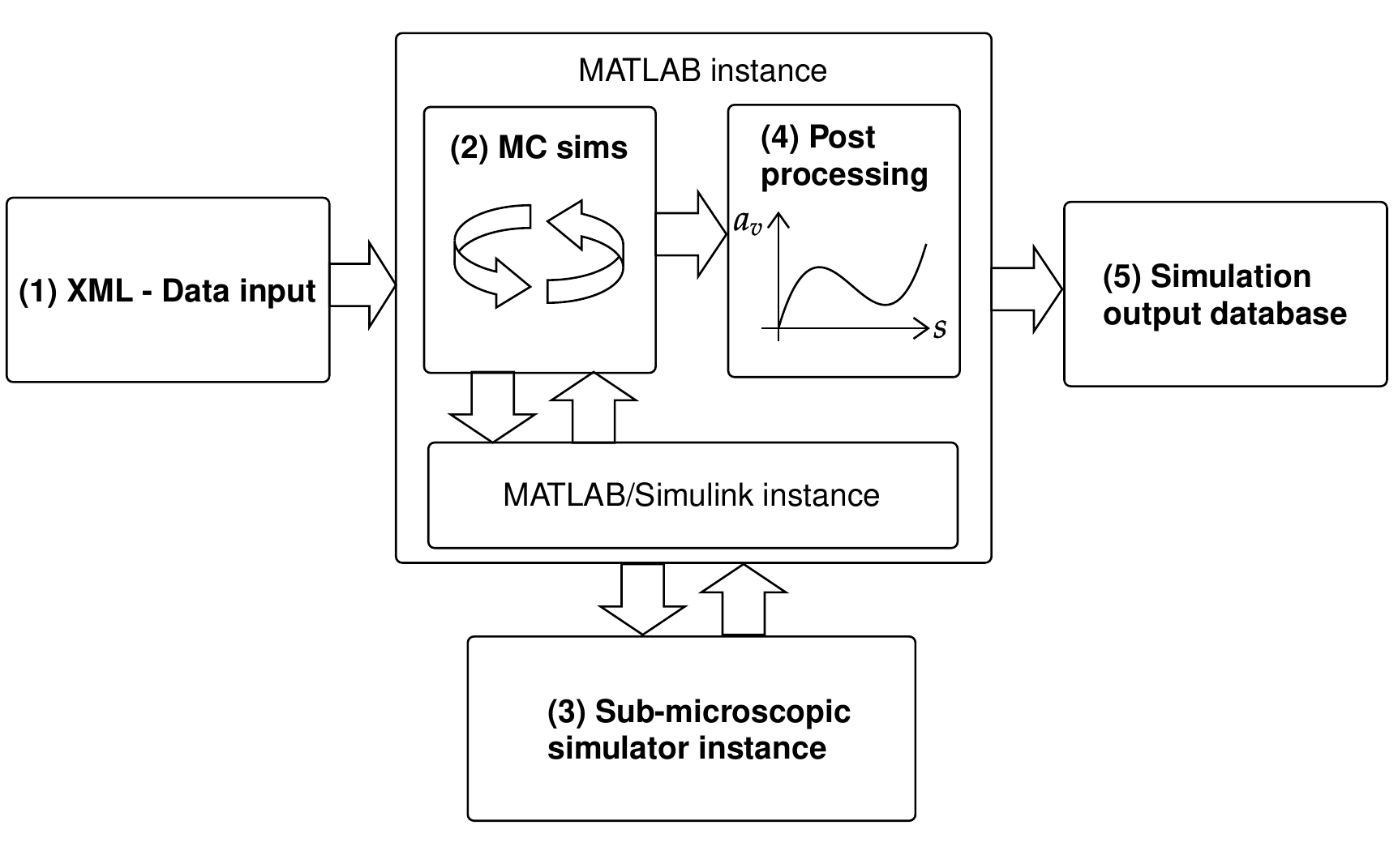}
    \caption{Simulation framework architecture.}
    \label{fig:framework}
\end{figure}

The presented framework provides a flexible tool that can be utilized to determine ride comfort data. Figure~\ref{fig:framework} depicts the modular simulation framework and all its components. The inputs for the framework are configured in an XML-file. This provides maximal flexibility when specifying test scenarios and the corresponding probabilistic input parameters (block~(1)~in~Fig.~\ref{fig:framework}). The core is implemented with MATLAB/Simulink and provides the automation functionality, i.e.\ the MC/LHS techniques (see Section~\ref{sec:MC} for details), and the communication interface to the sub-microscopic simulation environment (block~(2)~in~Fig.~\ref{fig:framework}). CarMaker from IPG Automotive is used as a simulation tool~\citep{ref:IPG}, which offers the possibility to simulate AVs that computes a driving velocity based on an speed profile or utilizes an ACC. Also, a motion planning algorithm is incorporated that provides motion profiles based on the configured driving style.  Here the interface between MATLAB/Simulink and CarMaker is utilized for the set-up, configuration of simulation inputs, as well as for exporting the output data (block~(3)~in~Fig.~\ref{fig:framework}). In block~(4)~(Fig.~\ref{fig:framework}), a database stores the output data, which are used for the ride comfort calculation. Here the methodologies for ride comfort estimation (i.e., thresholding, ISO 2631, and IRI) are applied. Finally, the ride comfort data are stored again in a database (block~(5)~in~Fig.~\ref{fig:framework}).

Due to the generic design of the framework, input or output blocks can be readily replaced by different representation technologies. Furthermore, the simulation environment CarMaker can be replaced by any other simulator, compatible with MATLAB/Simulink and OpenCRG, as explained in the next section. In addition, the framework allows for an extension to Software in the loop (SIL) or Hardware in the loop (HIL) tests. Note, that such changes would not deteriorate the performance of the developed framework.

\subsection{Road surface modeling}
\label{sec:road_surf}
Given the objective to determine high-resolution ride comfort estimates, the simulation output should represent the real-world interactions between road infrastructure and vehicles as accurately as possible. Because the default CarMaker settings for replicating a road surface offer limited complexity, the modeling standard OpenCRG is used in the current implementation. OpenCRG uses a 2D Curved Regular Grid (CRG) for representing elevation data based on a proximal reference line. To derive a road surface model (i.e., road profile and reference line) the following empirical steps are necessary: Using a mobile mapping vehicle, data from a profile laser scanner are collected (900 points per profile over a transversal width of 4m, 800 profiles/sec.) together with a precise trajectory of the vehicle (sampling frequency 200 Hz). After post-processing the trajectory using a static base station, a point cloud is derived from the laser scanner's raw data and trajectory. From the trajectory, a Catmull–Rom spline is interpolated to be used as reference line. Consequently, the reference line is specified with a start/end location as well as heading angles. Along the reference line, equidistant transversal profiles are calculated with a sampling distance of $0.1{\times}0.1$ m~\cite{ref:opencrg}.

ASCII format is used for representing a CRG-file containing post-processed point data from a laser scanner. CarMaker allows importing such road surface files and interprets them according to the standard. For robust surface integration, an error detection procedure (i.e.,\ measurement failures, point data outliers, etc.) must be implemented. The simulation software provides different interpolation methods. For this work, a cubic spline interpolation in the directions $x$, $y$, and $z$ is utilized. For deriving an accurate road representation in the simulation, a spline smoothing is performed to define the best parameters $\lambda_x$, $\lambda_y$, and $\lambda_z$ that minimize the total error of the model.

\subsection{Latin Hypercube Sampling technique}
\label{sec:MC}
By default, a simulation run is performed with an AV following an ideal trajectory, i.e.,\ in the middle of a road lane segment and a predefined speed profile. Essentially, this does not lead to representative vehicle dynamics that would allow for a reasonable ride comfort determination, as driving styles and speed choices usually diverge. Consequently, stochastic input parameters are introduced. All simulation runs are fed with random input samples to introduce  a deviation from the idealized set-up. Consequently, a representative data set for every road section under investigation is collected. To reduce the computational complexity, the LHS technique is employed. The method follows the procedure of dividing the Probabilistic Density Function (PDF) of an input into $n$ non-overlapping intervals, where each interval represents an equal probability value. Samples are generated in random order from every $n$ with respect to the corresponding interval density value. Dependent on the number of input variables used for the sampling, denoted as $m$, an ($n{\times}m$) matrix can be created~\cite{ref:LHS}. Note that the use of LHS requires a representative number of samples to obtain statistically meaningful simulation results. Different works in the literature (see e.g.,~\cite{ref:LHS_rep}) propose guidelines on guaranteeing a specific confidence interval. 
In our work, we denote the deviation from a given speed profile sample $v_{\mathrm{dev}}$ as a normal distribution with a mean value $\mu_{\mathrm{dev}}$ and standard deviation $\sigma_{\mathrm{dev}}$; i.e., $\mathcal{N}_{v_{\mathrm{dev}}} \sim \mathcal{N}(\mu_{\mathrm{dev}}, \sigma_{\mathrm{dev}})$. For the lateral lane position, we follow the same approach and introduce the lateral deviation from the ideal vehicle trajectory as $l_p$; hence, we denote the PDF as $\mathcal{N}_{l_{\mathrm{p}}} \sim \mathcal{N}(\mu_{\mathrm{p}}, \sigma_{\mathrm{p}})$. To additionally model weather conditions, we utilize the friction coefficient of the road surface $\mu_{\mathrm{rs}}$  as a proxy. The input values are sampled from a uniform distribution denoted by $\mathcal{U}_{\mathrm{rs}} \sim \mathcal{U}(a_{\mathrm{rs}},b_{\mathrm{rs}})$, where $a_{\mathrm{rs}}$ and $b_{\mathrm{rs}}$ are the distribution parameters.    

\section{VEHICLE MODEL VALIDATION AND PARAMETERS OPTIMIZATION}
\label{sec:model_val_opti}
CarMaker software provides state-of-the-art modeling for the sub-microscopic simulation environment of virtual test driving. Although its models have been tested extensively, validated output data is not guaranteed. Because ride comfort estimation depends heavily on the simulated vehicle dynamics, a model validation procedure has been applied to determine the respective error. Therefore, a ground-truth data pool of vehicle dynamics should be available for comparison, such as, e.g., measurements by an Inertial Measurement Unit (IMU). For the validation, the velocity in the $x$-direction $v_x$, accelerations in the three-dimensional space $a_x$, $a_y$, and $a_z$ and the Euler angle rates, i.e., roll $\Phi$-, pitch $\Theta$- and yaw $\Psi$-rate, are utilized. All parameters with their corresponding units are listed in Table~\ref{tab:quantities}.

\begin{table}[t]
	\centering
	\small
	\caption{Vehicle dynamics quantities for model validation and optimization.}
\begin{tabular}{lll}
	\toprule
	Variable     & Unit        & Description                                 \\
	\midrule
	T         & {[}s{]}     & Time vector of test run/measurement                     \\
	$v_x$        & {[}m/s{]}     & Absolute velocity of vehicle body in x-direction        \\
	$a_x$       & {[}m/s$^2${]}  & Acceleration of vehicle body in x-direction \\
	$a_y$       & {[}m/s$^2${]}  & Acceleration of vehicle body in y-direction \\
	$a_z$       & {[}m/s$^2${]}  & Acceleration of vehicle body in z-direction \\
	$\Phi$-rate  & {[}deg/s{]} & Roll-rate of vehicle body                   \\
	$\Theta$-rate & {[}deg/s{]} & Pitch-rate of vehicle body                  \\
	$\Psi$-rate  & {[}deg/s{]} & Yaw-rate of vehicle body   \\
	\bottomrule                
\end{tabular}
\label{tab:quantities}
\end{table}

\begin{figure}[!b]
	\centering
	\includegraphics[width=1.0\textwidth]{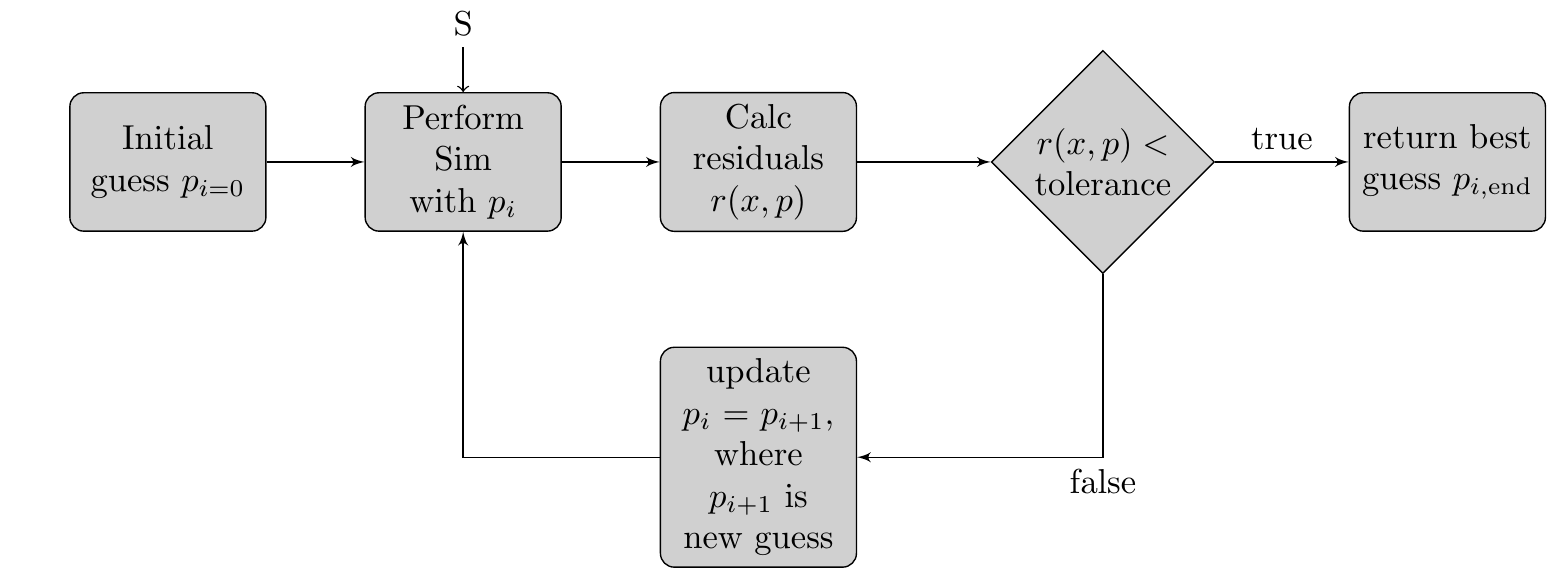}
	\caption{Flowchart of non-linear least square optimization.}
	\label{fig:leastsquare}
\end{figure}

Essentially, the real-world data can be used to evaluate the performance of the simulation. To assess the error of all quantities in Table~\ref{tab:quantities}, performance metrics such as Root Mean Square Error (RMSE) are commonly used. Nevertheless, the quantities in Table~\ref{tab:quantities} have different physical definitions and magnitudes. Hence, we normalize RMSE to the Normalized RMSE (NRMSE). The residuals can then be combined and formulated as follows:
\begin{equation}
	\label{eq:nonlinlsq}
	 r(x,p) = f(x,p) - y(x),
\end{equation} 
where the function $f(x,p)$ defines the non-linear simulation model with vectors $x$ and $p$ denoting the input and model parameter sets, respectively; $y(x)$ denotes the measurement data (e.g., from an IMU device) and $r(x,p)$ the corresponding residuals with respect to $x$ and $p$. 
We can now utilize $r(x,p)$ (by taking~(\ref{eq:nonlinlsq})) as our objective function to formulate a non-linear simulation-based optimization problem, aiming to minimize the residuals. We start describing the optimization procedure to minimize $r(x,p)$ by explaining the algorithm steps depicted in Fig.~\ref{fig:leastsquare}. 
Initially, a simulation is performed with an initial guess of vector $p$, i.e., $p_{i=0}$. Based on the result $r(x,p)$, the non-linear optimization Levenberg-Marquardt algorithm from~\cite{ref:Levenberg} is initialized. For every iteration, the residuals are checked against a certain objective tolerance. When the result does not satisfy the postulated accuracy, $p_i$ is updated with $p_{i+1}$, which is the new gradient-based guess. When the objective function's magnitude falls below the desired tolerance level, a solution $p_{i,{\rm end}}$ is returned. Note that the input $S$ represents the simulation scenario's modeling, e.g., test vehicle, road surface, etc., which does not change over time/runs.

\begin{figure}[tb]
	\centering
	\includegraphics[width=1.0\textwidth]{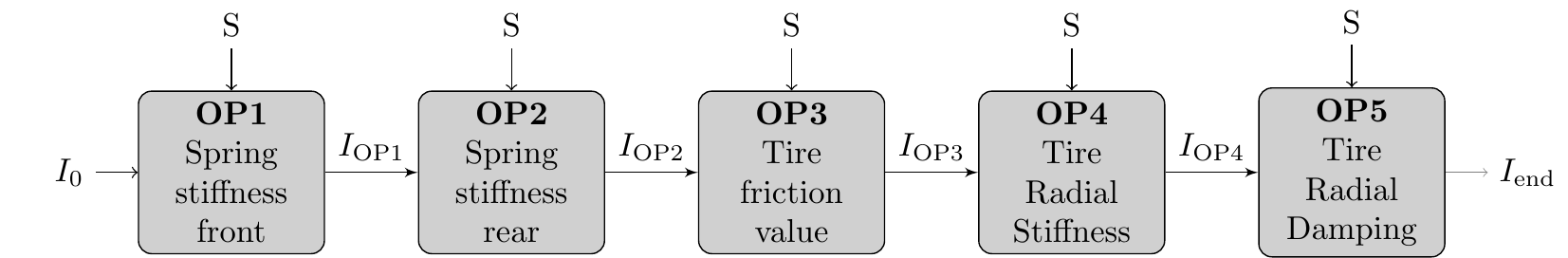}
	\caption{Optimization procedure, where every block represents a non-linear least square optimization.}
	\label{fig:optichain}
\end{figure}

The error is minimized by an optimization procedure described in Fig.~\ref{fig:optichain}. We select five parameters based on a sensitivity analysis from~\cite{ref:bmw}. The parameter values for this work were set based on their (a) effectiveness shown by~\cite{ref:bmw} (b) availability and (c) accessibility in the simulator.
Five optimization steps are computed. The signals $I_{\rm 0}$, $I_{{\rm end}}$ and $I_{{\rm OP}_i}$ in Fig.~\ref{fig:optichain} are the initial input, output of the optimization procedure, and preliminary results between two optimization blocks ${\rm OP}_i$ and ${\rm OP}_{i-1}$, with $i>0$, respectively. By defining reasonable lower and upper bounds for the parameters set, the optimization problem can be formulated as follows:
\begin{equation}
\label{eq:opti}
\begin{aligned}
&\underset{p \in \!R}{\text{minimize}}
& & r(x,p) \\
& \text{subject to:}
& & K_{\rm sf,\min} \leq K_{\rm sf} \leq K_{\rm sf,\max} \\
& & & K_{\rm rf,\min} \leq K_{\rm rf} \leq K_{\rm rf,\max} \\
& & & \mu_{\rm Tr,\min} \leq \mu_{\rm Tr} \leq \mu_{\rm Tr,\max} \\
& & & K_{\rm Tr,\min} \leq K_{\rm Tr} \leq K_{\rm Tr,\max} \\
& & & d_{\rm Tr,\min} \leq d_{\rm Tr} \leq d_{\rm Tr,\max}, 
\end{aligned}
\end{equation}
where $K_{\rm sf}$ and $K_{\rm rf}$ denote the front and rear spring stiffness, respectively, $\mu_{\rm Tr}$ the value of tire friction, $K_{\rm Tr}$ the tire radial stiffness, and $d_{\rm Tr}$ the tire radial damping. The constraints define the corresponding minimum and maximum values for all parameters. The optimization problem in~(\ref{eq:opti}) aims at computing the best parameter values, so as to minimize the error between the simulator and the real data. The obtained set of parameters is then applied to the simulation framework presented in Section~\ref{sec:sim_framework} and allows the derivation of high accuracy comfort data. The methodologies for comfort estimation are discussed in the next section.

\section{RIDE COMFORT DETERMINATION STRATEGY}
\label{sec:ride_comfort}
To stress the importance of a robust ride comfort classification, we have based the determination of critical ride comfort on three common strategies in this work. First, established threshold values for vehicle accelerations are utilized to classify comfort. The compilation of different acceleration bands and the corresponding technical application is described in Section~\ref{sec:thresholding}. Second, this work considers the commonly used ISO 2631 guideline, allowing for a comfort evaluation by deriving a frequency-weighted acceleration signal. Section~\ref{sec:iso_comfort} explains the procedure in detail. Finally, we compare the two evaluation strategies with empirical IRI data, allowing for comfort classification. The concept of the IRI is introduced in Section~\ref{sec:IRI_concept}. 

\subsection{Comfort classification with thresholding}
\label{sec:thresholding}
The application of threshold values in signal processing is a simple standard methodology to mitigate system exposure to extreme signal magnitudes. In the context of AVs, thresholding is applied (among others) to acceleration signals to ensure a realistic system behavior and a comfortable user experience. In this work, we utilize threshold values to determine comfort based on an extensive collection by~\cite{ref:bae_threshold_comfort}. This work collects comfort thresholds for public transportation (PT) and two different driving styles, i.e., normal driving (ND) and aggressive driving (AD) for the longitudinal and lateral acceleration, respectively. As the driver does not actively influence the AVs' motion planning, public transportation thresholds are of great interest and logically included in the analysis.
In~\cite{ref:bae_threshold_comfort} no threshold values for the acceleration and deceleration in the z-axis; $a_z$ and $-a_z$ (i.e., the vertical plane) are mentioned. To the authors' knowledge, no commonly used threshold for such quantities is available. Hence, the authors investigated a data set from~\cite{ref:michalis_avs}, where the impacts of ACC on traffic flow and string stability are studied. Results show that a lower and upper border for $a_z$ and $-a_z$ can be derived at $0.1$m/s$^2$ and $-0.1$m/s$^2$, respectively. As~\cite{ref:michalis_avs} only investigate vehicle maneuvers that are classified as normal driving, this work investigates the vertical plane with an ND threshold. For the PT thresholds in the z-axis, magnitude is assumed to be equal to ND. For AD, we triple the magnitude and set the thresholds to -0.3m/s$^2$ for the deceleration and 0.3m/s$^2$ for the acceleration.  Table~\ref{tab:thresholds_manual} lists the utilized values for longitudinal acceleration $a_x$, deceleration $-a_x$, lateral acceleration $a_y$ and deceleration $-a_y$, and vertical acceleration $a_z$ and deceleration $-a_z$, respectively. 

\begin{table}[b]
  \centering
  \small
  \caption{Utilized threshold values for comfort determination in the longitudinal (denoted with the x-axis), lateral (denoted with the y-axis), and vertical direction (denoted with the z-axis). Values are based on~\cite{ref:bae_threshold_comfort} and a data set analysis from~\cite{ref:michalis_avs}.}
  \begin{tabular}{lcc|lcc|lcc}
  \toprule
     ($x$-axis) & $-a_x$ & $a_x$ & ($y$-axis) & $-a_y$ & $a_y$ & ($z$-axis) & $-a_z$ & $a_z$ \\
     \midrule
    PT    & -0.90  & 0.90   & PT    & -0.90  & 0.90   & PT    & -0.10  & 0.10 \\
    ND    & -2.00    & 1.47  & ND    & -4.00    & 4.00     & ND    & -0.10  & 0.10 \\
    AG    & -5.08 & 3.07  & AG    & -5.60  & 5.60   & AG    & -0.30 & 0.30 \\
    \bottomrule
    \end{tabular}%
  \label{tab:thresholds_manual}%
\end{table}%

The thresholds are applied in the following to acceleration signal data to determine critical ride comfort roadway sections (Section~\ref{sec:th_methdology}).    

\subsection{ISO 2631 comfort classification}
\label{sec:iso_comfort}
To provide an alternative classification for ride comfort concerning certain frequency ranges of acceleration signals, investigations based on ISO 2631 are performed. Generally speaking, low-frequency contents of a signal, i.e.,\ for health, comfort, and perception between 0.5Hz and 80Hz, and for motion sickness ranging from 0.1Hz to 0.5Hz are considered. A frequency weighting technique based on ISO-guideline is used to examine the different applications and frequency ranges. The frequency weights are chosen upon the passengers' position (i.e.,\ standing, sitting, or recumbent) and the application (i.e.,\ human health and comfort, probability of vibration perception or occurrence of motion sickness). The weighting factor is chosen upon the application domain and indicated with the subscript $b$ from ISO 2631, where $b \in \{c, d, e, f, j, k\}$. Depending on the situation one need to investigate (e.g., comfort for a seated person by utilizing acceleration $a_x$) the subscript influences the properties of the signal processing presented in the following.

Let $a_{w_b}$ be the frequency-weighted acceleration signal by the type $b$. In order to obtain $a_{w_b}$, a filtering procedure with the corresponding filter design has to be performed. A high pass $H_h(p)$, low pass $H_l(p)$, acceleration-velocity transition $H_t(p)$, and an upward step filter $H_s(p)$ are introduced in~\cite{ref:ISO2631}. The filter equations are defined in the continuous time domain and given by:
\begin{equation}
	\label{eq:highpass}
	H_h(p) = \left\lvert \frac{1}{1+ \frac{\sqrt{2}  \omega_1}{p} + \frac{\omega_1^2}{p^2}} \right\rvert,
\end{equation}

\begin{equation}
	\label{eq:lowpass}
	H_l(p) = \left\lvert \frac{1}{1+ \frac{\sqrt{2}  p}{\omega_2} + \frac{p^2}{\omega_2^2}} \right\rvert ,
\end{equation}

\begin{equation}
	\label{eq:filter_ht}
	H_t(p) = \left\lvert \frac{1+\frac{p}{\omega_3}}{1+\frac{p}{Q_4\omega_4} + \frac{p^2}{\omega_4^2}} \right\rvert ,
\end{equation}

\begin{equation}
\label{eq:h_s}
	H_s(p) = \left\lvert \frac{1+\frac{p}{Q_5\omega_5} + \frac{p^2}{\omega_5^2}}{1+\frac{p}{Q_6\omega_6 + \frac{p^2}{\omega_6^2}}}\frac{\omega_5^2}{\omega_6^2}  \right\rvert ,
\end{equation}
where $\omega_i = 2 \pi f_i$; $Q_i$ and $f_i$ are the inputs for the transfer functions and defined as the resonant quality factors and corner frequencies, respectively; the index $i \in \{1,2,3,4,5,6\}$; for further details the reader is referred to the guidelines in~\cite{ref:ISO2631}. Equations~(\ref{eq:highpass})--(\ref{eq:h_s}) represent the filter equations in the continuous time domain $p$. For computational reasons, a conversion to the discrete time domain $z$ is performed by means of a bi-linear transformation:
 
\begin{equation}
	\label{eq:transf}
	H(z) = H(p) =  H\left(\frac{2(1-z^{-1})}{(1+z^{-1})}\right),
\end{equation}
where $H(z)$ denotes the total weighting function. By transforming $H(p)$ in the digital domain $z$, which is obtained by replacing $p = (2(1-z^{-1}))/(1+z^{-1})$ in equations~(\ref{eq:highpass})--(\ref{eq:h_s}), a digital filter can be designed~\citep{ref:dyer}. A full list of the rearranged terms required to fulfill the requested form of the filter design can be found in~\cite{ref:rimell}.
Finally, the filtering of the acceleration signal can be applied according to the following equation:
\begin{equation}
\label{eq:eva_transfer}
	H(z) = H_h(z)  H_l(z)  H_t(z)  H_s(z).
\end{equation}
Note that there are exceptions where certain filter operators of equation~(\ref{eq:eva_transfer}) are set to $H_o(z) = 1$, where $o \in \{h,l,t,s\}$; this again depends on the application. For further details the interested reader is referred to~\cite{ref:ISO2631}.

The filtered output signal is evaluated by computing the Root Mean Square (RMS) value, defined as follows:
\begin{equation}
\label{eq:rms_acc}
    a_{{w_b},\rm RMS} =  \sqrt{\frac{1}{T}\int_{0}^{T} a_{w_b}^2(t) dt},
\end{equation}
where $a_{{w_b},\rm RMS}$ is the RMS value of $a_{w_b}$ and $T$ the duration of measurements for $a_{w_b}$. For the evaluation of comfort, ISO 2631 recommends the separate evaluation of directions $x$, $y$, and $z$, and the combination of the frequency weighted signals according to:
\begin{equation}
    \label{eq:combi}
	a_v = \sqrt{k_x^2a^2_{{w_{b,x},\rm RMS}} + k_y^2a^2_{{w_{b,y},\rm RMS}} + k_z^2a^2_{{w_{b,z},\rm RMS}}}.
\end{equation}
Here, $a_v$ represents the total vibration value, $a_{w_{b,x}, \rm RMS}$, $a_{w_{b,y}, \rm RMS}$, and $a_{w_{b,z}, \rm RMS}$ the frequency weighted RMS acceleration signals, and $k_x = k_y = k_z = 1$ the corresponding ISO 2631 weighting factors per direction, respectively. The values of $k_x$, $k_y$, and $k_z$ diverge from the proposed values only in special designs, i.e.,\ a seated person, affected by vibrations (a) of the backrest, (b) at the feet, or (c) around a rotation axis. The post-processed acceleration signal $a_v$ allows for the comfort level determination, according to the recommended threshold levels from~\cite{ref:ISO2631} listed in Table~\ref{tab:acc_thresholds}. Consequently, the thresholding allows a classification into levels ranging from not uncomfortable (NU) up to extremely uncomfortable (EU). 

\begin{table}[b]
  \centering
  \small
  \caption{Threshold levels for $a_v$.}
    \begin{tabular}{ll}
    \toprule
    Acceleration magnitude & Uncomfort Level \\
    \midrule
    $a_v <$ 0.315 m/s$^2$ & Not uncomfortable (NU) \\
    0.315 m/s$^2$ $< a_v <$ 0.63 m/s$^2$ & Little uncomfortable (LU) \\
    0.50 m/s$^2$ $< a_v <$ 1 m/s$^2$ & Fairly uncomfortable (FU) \\
    1.25 m/s$^2$ $< a_v <$ 2.50 m/s$^2$ & Very uncomfortable (VU) \\
    $a_v$ $> 2 m/s^2$ & Extremely uncomfortable (EU) \\
    \bottomrule
    \end{tabular}%
  \label{tab:acc_thresholds}%
\end{table}%

In addition, the guidelines recommend a threshold range for the probability of vibration perception from 0.01m/s$^2$ to 0.02m/s$^2$. Investigations on human health and motion sickness have been neglected in this study, as vibrations have to be present for a certain amount of time and with a certain magnitude (e.g.,\ negative impact on human health at $\sim$1m/s$^2$ for four hours~\cite{ref:ISO2631}), which is not common in vehicle dynamics. Finally, the processed signal $a_v$ is utilized to determine critical ride comfort roadway sections with the proposed methodology in Section~\ref{sec:th_methdology}. 

\subsection{IRI comfort classification}
\label{sec:IRI_concept}
Development of the International Roughness Index (IRI) started in the late 1960s and went through several stages of development after it was finally introduced by the World Bank as it is known today. IRI constitutes a dimensionless roughness index of a longitudinal road profile and is widely used in pavement assessments but is also deemed suitable to determine ride comfort. 

To derive a robust and reproducible IRI, the methodology went through several stages of development. In the 1960s, researchers already choose a quarter-car-model to simulate the interaction between a vehicle and the road surface. As this model involves several parameters, a calibration by correlation procedure was carried out by processing data from the International Road Roughness Experiment (IRRE)~\citep{ref:gillespie}. Consequently, the well-known Golden Car parameters were derived for a half- and quarter-car-model. As both models showed similar correlation level and the quarter-car-model allows usage with all profiling methods, today's applications are still based on the quarter-car-model. Also, researches widely agree that IRI should be measured at a velocity of 80 km/h~\cite{ref:mucka_iri_2020}. 

To define the quarter-car-model (definition based on~\cite{ref:sayers_IRI_1995}) we assume a system consisting of a sprung mass (i.e., vehicle body mass supported by one wheel) denoted as $m_{\mathrm{s}}$ and an unsprung mass (i.e., wheel or tire) denoted as $m_{\mathrm{u}}$; $m_{\mathrm{s}}$ and $m_{\mathrm{u}}$ are connected through a suspension and a damper. For the quarter-car-model, the suspension spring rate $k_{\mathrm{s}}$ and suspension damping rate $c_{\mathrm{s}}$ are considered. Further, $k_{\mathrm{t}}$ denotes the tire spring rate. As mentioned above, the Golden Car parameters derived during the model calibration process are utilized:  
$c = c_{\mathrm{s}}/m_{\mathrm{s}} = 6.00$; $k_1 = k_t/m_{\mathrm{s}} = 653.00$; $k_2 = k_s/m_{\mathrm{s}} = 63.30$; $\mu = m_{\mathrm{u}}/m_{\mathrm{s}} = 0.15$. Note that for simplicity all parameters are normalized by sprung mass $m_{\mathrm{s}}$. 

Finally, the quarter-car-model defined by first-order ordinary differential equations (ODE) can be written in compact form as: 
\begin{equation}
        \Dot{x} = Ax + Bh_{\mathrm{ps}},
\end{equation}
where vector $\Dot{x}$ and matrices $A$ and $B$ are denoted as 
\begin{equation}
   x =  \begin{bmatrix}
        z_{\mathrm{s}} & \Dot{z_{\mathrm{s}}} & z_{\mathrm{u}} & \Dot{z_{\mathrm{u}}}
\end{bmatrix}^T,
\end{equation}
\begin{equation}
   A =  \begin{bmatrix}
        1 & 0 & 0 & 0\\
        -k_2 & -c & -k_2 & c\\
        0 & 0 & 1 & 0\\
        \frac{k_2}{\mu} & \frac{c}{\mu} & \frac{k_1 + k_2}{\mu} & -\frac{c}{\mu}
\end{bmatrix},
\end{equation}
and 
\begin{equation}
   B =  \begin{bmatrix}
       0 & 0 & 0 & \frac{k_1}{\mu}
\end{bmatrix}^T.
\end{equation}
$h_{\mathrm{ps}}$ denotes the smoothed elevation profile; $z_{\mathrm{s}}$ and $z_{\mathrm{u}}$ the vertical height of the sprung and unsprung mass, respectively; $\bold{x}$ the state vector holding  $z_{\mathrm{s}}$ and $z_{\mathrm{u}}$ with its corresponding time derivatives $\Dot{z_{\mathrm{s}}}$ and $\Dot{z_{\mathrm{u}}}$, respectively. 
As IRI can be computed by accumulating the absolute difference of derivatives $\Dot{z_{\mathrm{s}}}$ and $\Dot{z_{\mathrm{u}}}$, a standard algorithm for solving ODEs needs to be applied as suggested in~\cite{ref:sayers_IRI_1995}. Finally, the IRI is defined as follows:

\begin{equation}
    \mathrm{IRI} = \frac{1}{L} \int_0^{L/V} | \Dot{z_{\mathrm{s}}} - \Dot{z_{\mathrm{u}}} | dt,
\end{equation}
where $L$ is the profile length and $V$ is the measurement speed. 

The calculated index can be used to quantify road roughness. Therefore, thresholds allowing for classification of ride quality ratings into very good, good, fair, mediocre, and poor are defined by~\cite{ref:FHWA_RC_thresholds}. Nevertheless, for classification, the assumption of a constant vehicle speed of $V=80$ km/h must be full-filled. To overcome this limitation,~\cite{ref:Yu_IRI_comfort_THs} defines speed-dependent IRI-thresholds (see Table~\ref{tab:IRI_thresholds}). 

\begin{table}[t]
  \centering
  \small
  \caption{IRI thresholds as a function of speed~\cite{ref:Yu_IRI_comfort_THs}.}
    \begin{tabular}{llllll}
    \toprule
          & \multicolumn{5}{c}{IRI Threshold at different speeds (m/km)} \\
    Ride quality & 120 (km/h) & 100 (km/h) & 80 (km/h) & 70 (km/h) & 60 (km/h)  \\
    \midrule
    Very good (VG)  & $<$ 0.95 & $<$1.14 & $<$1.43 & $<$1.63 & $<$1.90 \\
    Good (G)  & 0.95–1.49 &  1.14–1.79 &  1.43–2.24 &  1.63–2.57 &  1.90–2.99 \\
    Fair (F)  & 1.50–1.89  & 1.80–2.27 &  2.25–2.84 &  2.58–3.25 &  3.00–3.79 \\
    Mediocre (M) & 1.90–2.70 &  2.28–3.24 &  2.85–4.05  & 3.26–4.63 &  3.80–5.40 \\
    Poor (P)   & $>$2.70 &  $>$3.24 &  $>$4.05 & $>$4.63 &  $>$5.40 \\
      \midrule
    Ride quality & 50 (km/h) & 40 (km/h) & 30 (km/h) & 20 (km/h) & 10 (km/h) \\
      \midrule
    Very good (VG)  & $<$2.28 & $<$2.86 & $<$3.80 & $<$5.72 & $<$11.44 \\
    Good (G) & 2.28–3.59 &  2.86–4.49  & 3.80–5.99 &  5.72–8.99 &  11.44–17.99 \\
    Fair (F) & 3.60–4.54 &  4.50–5.69 &  6.00–7.59 &  9.00–11.39 &  18.00–22.79 \\
    Mediocre (M) & 4.55–6.25 &  5.70–8.08 &  7.60–10.80 &  11.40–16.16 &  22.80–32.32 \\
    Poor (P)  & $>$6.25 &  $>$8.08 &  $>$10.80 &  $>$16.16 &  $>$32.32 \\
    \bottomrule
    \end{tabular}%
  \label{tab:IRI_thresholds}%
\end{table}%
As expected, the threshold magnitudes for all categories of ride quality increase with decreasing speed. To derive comfort estimates, thresholds from Table~\ref{tab:IRI_thresholds} are applied dynamically to the IRI measurement by considering a given speed profile. Again, we utilize the final signal to derive critical comfort road sections (Section~\ref{sec:th_methdology}). 

\subsection{Derivation of critical ride comfort roadway sections}
\label{sec:th_methdology}
To provide a quantitative analysis of critical comfort and obtain such road sections, we define a sliding window procedure. Let $\mathrm{TS}_i$ be a test track with an index $i$. Assuming that on $\mathrm{TS}_i$ multiple test drives are performed, consequently sets of accelerations $\mathcal{A}$, $\mathcal{B}$, and $\mathcal{C}$ are derived; i.e., $a_x(t) \in \mathcal{A}$, $a_y(t) \in \mathcal{B}$, and $a_z(t) \in \mathcal{C}$ for all test drives. As all elements of $\mathcal{A}$, $\mathcal{B}$, and $\mathcal{C}$ contain different speed profiles, the position of the vehicle in space (denoted as $s$) at a specific time step $t$ is not equivalent in all signals. Hence, a classification based on the time-dependent signals would not yield to the desired classification result. To overcome this issue, we transform all time-dependent acceleration signals to the space-domain. We merge 
samples of elements from $\mathcal{A}$, $\mathcal{B}$, and $\mathcal{C}$ at a specific point in space $s$. Hence, we derive $a_x(s)$, $a_y(s)$, and $a_z(s)$, which are utilized to apply a sliding window of $w$, where $w$ represents the length of critical comfort road sections in meters. We perform a convolution of the acceleration signal and the corresponding thresholds from Tables~\ref{tab:thresholds_manual},~\ref{tab:acc_thresholds}, and~\ref{tab:IRI_thresholds}. The window is slid over the acceleration signals, and if for a complete window the applied threshold is exceeded, a critical comfort road section is found. All the simulation outputs from Section~\ref{sec:sim_framework} are post-processed with the proposed methodology. Moreover, results from a case study in Austria are presented in Section~\ref{sec:case_study}.

\section{CASE STUDY}
\label{sec:case_study}
The following case study discusses the application of the proposed framework. Demonstration scenarios are configured that consist of three test sites with available data measurements. First, we present the three test sites where measurement data were recorded. Afterwards, we set up the simulation framework and perform a first model validation, followed by the proposed model parameters optimization. Results of the improvements are presented in Section~\ref{sec:model_optim}. To utilize our simulation framework with the MC/LHS option for deriving the vehicle dynamics data, we present the configuration in Section~\ref{sec:stochastic_sim_set_up}. Finally, we compare the ride comfort determination strategies, i.e., the thresholding procedure, ISO 2631 approach, and IRI classification. Classification of critical ride comfort roadway sections is presented for all studied methodologies, and a conclusion about the respective performance is drawn (see Section~\ref{sec:ride_comfort_results}).  

\begin{figure*}
    \centering
  \subfloat[]{%
       \includegraphics[width=0.55\linewidth]{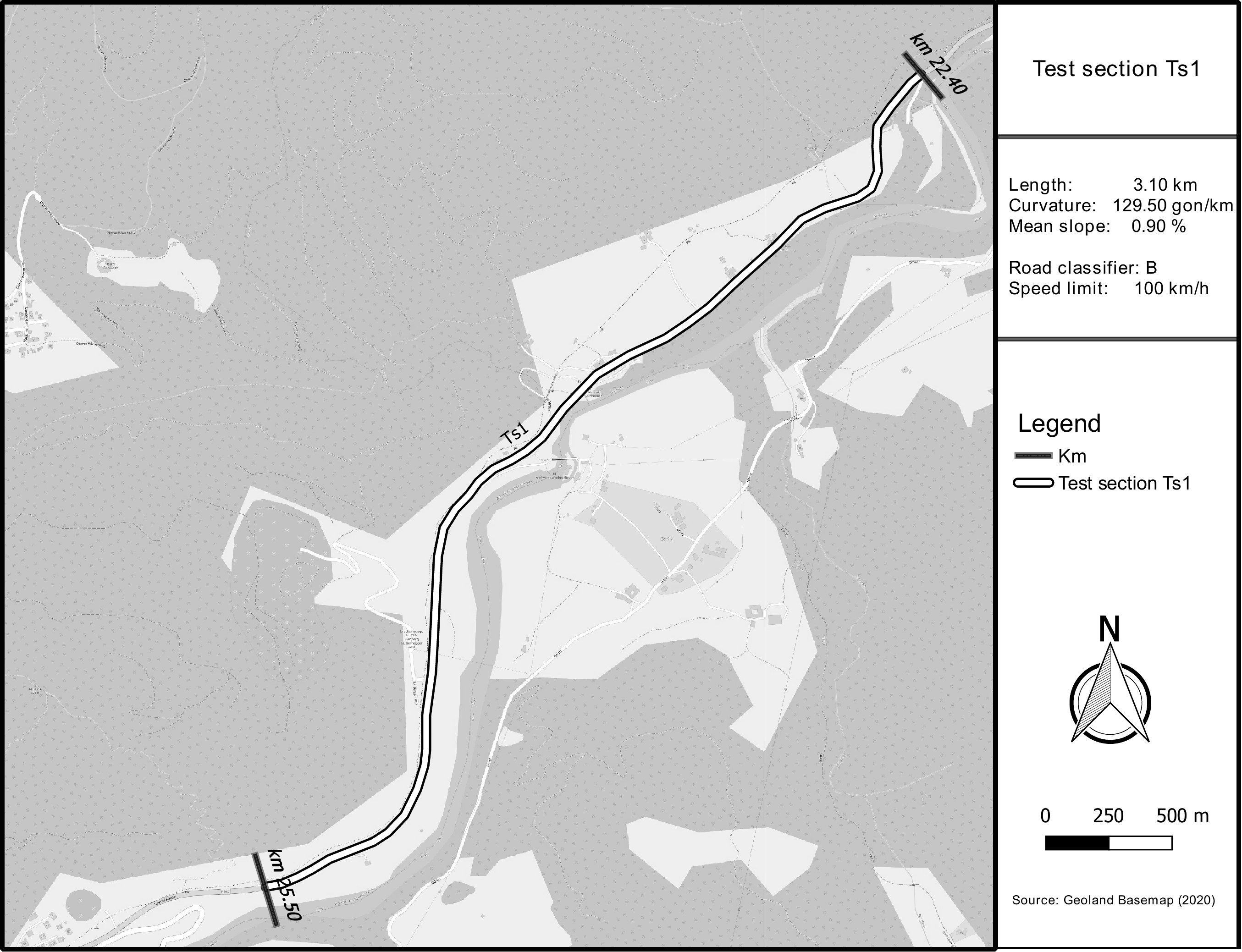}}
    \qquad
  \subfloat[]{%
        \includegraphics[width=0.55\linewidth]{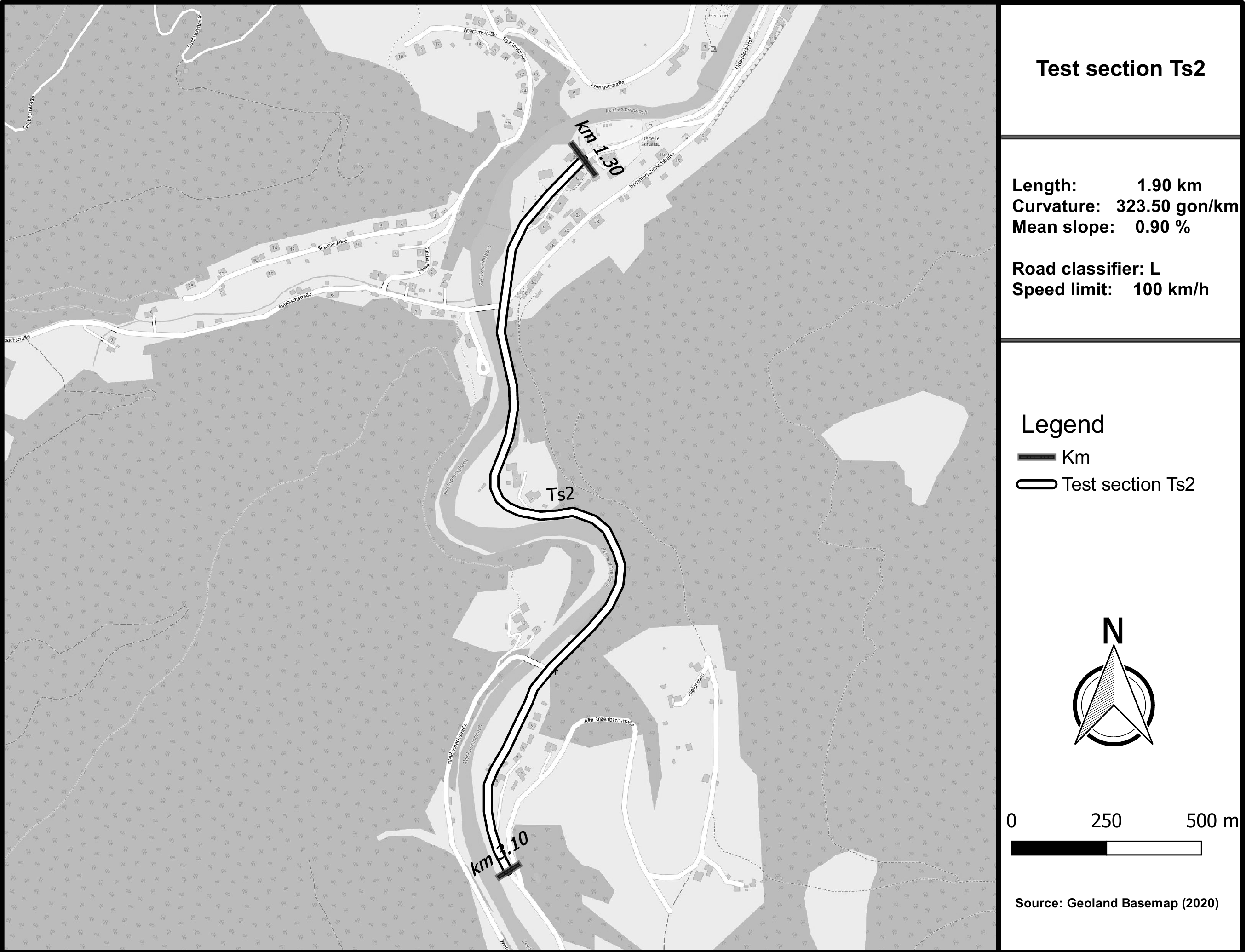}}
    \qquad
  \subfloat[]{%
        \includegraphics[width=0.55\linewidth]{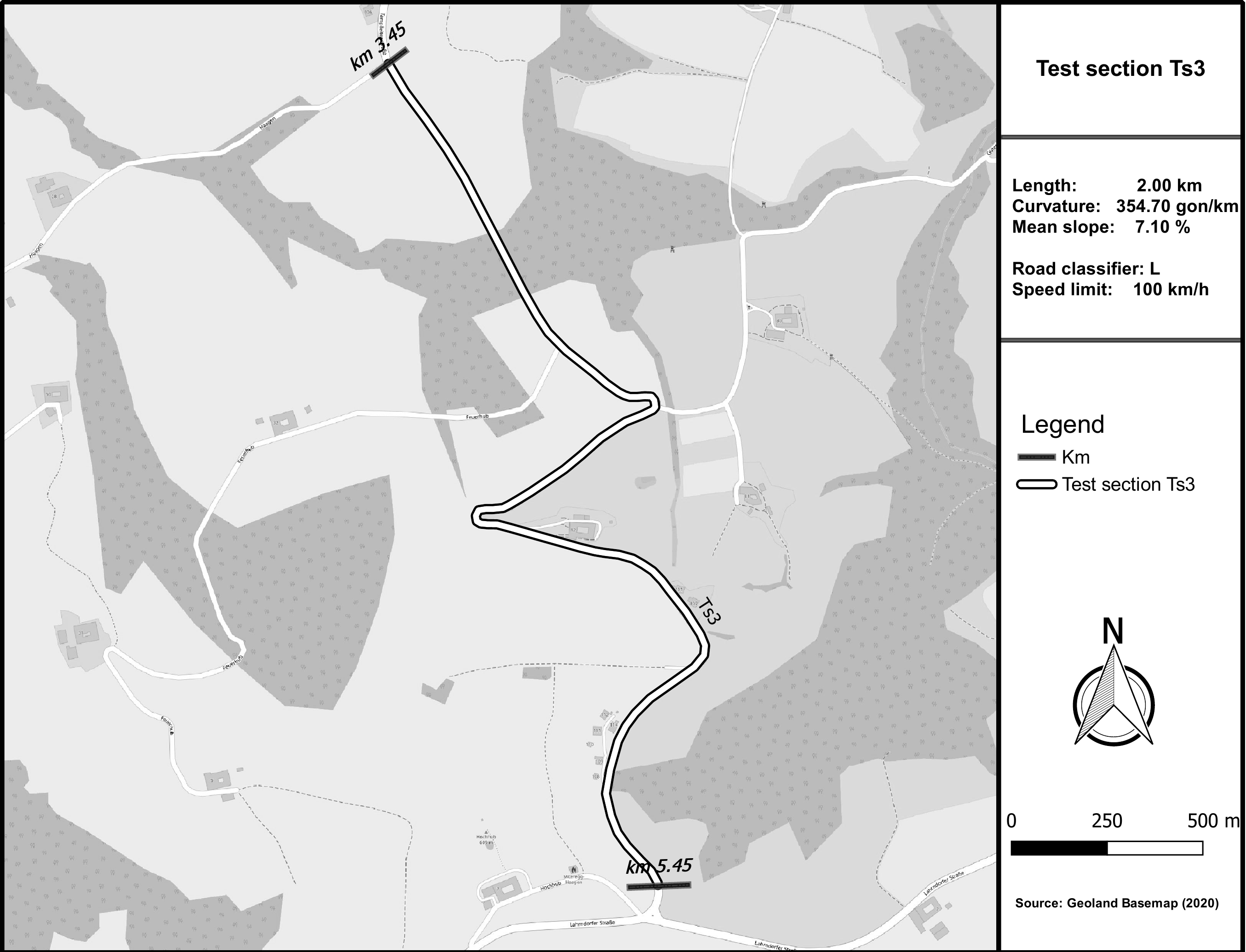}}
  \caption{Test sites characteristics: (a), (b) and (c) represent $Ts{\rm 1}$, $Ts{\rm 2}$ and $Ts{\rm 3}$, respectively.}
  \label{fig:testsection_figs} 
\end{figure*}

\subsection{Test sections description and measurements}
\label{sec:test_sec_meas}
Data have been collected from three test sites in Austria (see Fig.~\ref{fig:testsection_figs}) that differ in length, curvature, elevation profile, and average driving speed. $Ts{\rm 1}$ is a federal road (classifier B in Austria) with a length of 3.10 kilometers and a corresponding speed limit of 100 km/h. Average slope of the test section is 0.90\%. Second, $Ts{\rm 2}$ represents a minor road (classifier L in Austria) with a length of 1.90 kilometers and a corresponding speed limit of 100 km/h. The average slop is again 0.90\%. The third section is a minor road of 2.00 km length, denoted as $Ts{\rm 3}$. Its speed limit is again 100km/h, and average slope is 7.1\%. Figure~\ref{fig:testsection_figs} shows that the curvature of the test sections increase from $Ts{\rm 1}$ to $Ts{\rm 3}$; i.e., 129.50gon/km, 323.50gon/km and 354.70gon/km, respectively. Hence, the average driving speed per test section will decrease, although the speed limits are identical. The road characteristics of the studied sections provide a variety of data that exploits the robustness of both the optimization procedure and comfort methodologies. The measurements were performed with an IMU capable of tracking high-resolution vehicle dynamics, mounted on the top of the test vehicle, and connected to a central unit. Data of the parameters listed in Table~\ref{tab:quantities} have been collected and serve as an input for the model parameters optimization.

\subsection{Vehicle model validation and parameters optimization}
\label{sec:model_optim}
\begin{figure}[t]
    \centering
  \subfloat[]{%
       \includegraphics[width=0.33\linewidth]{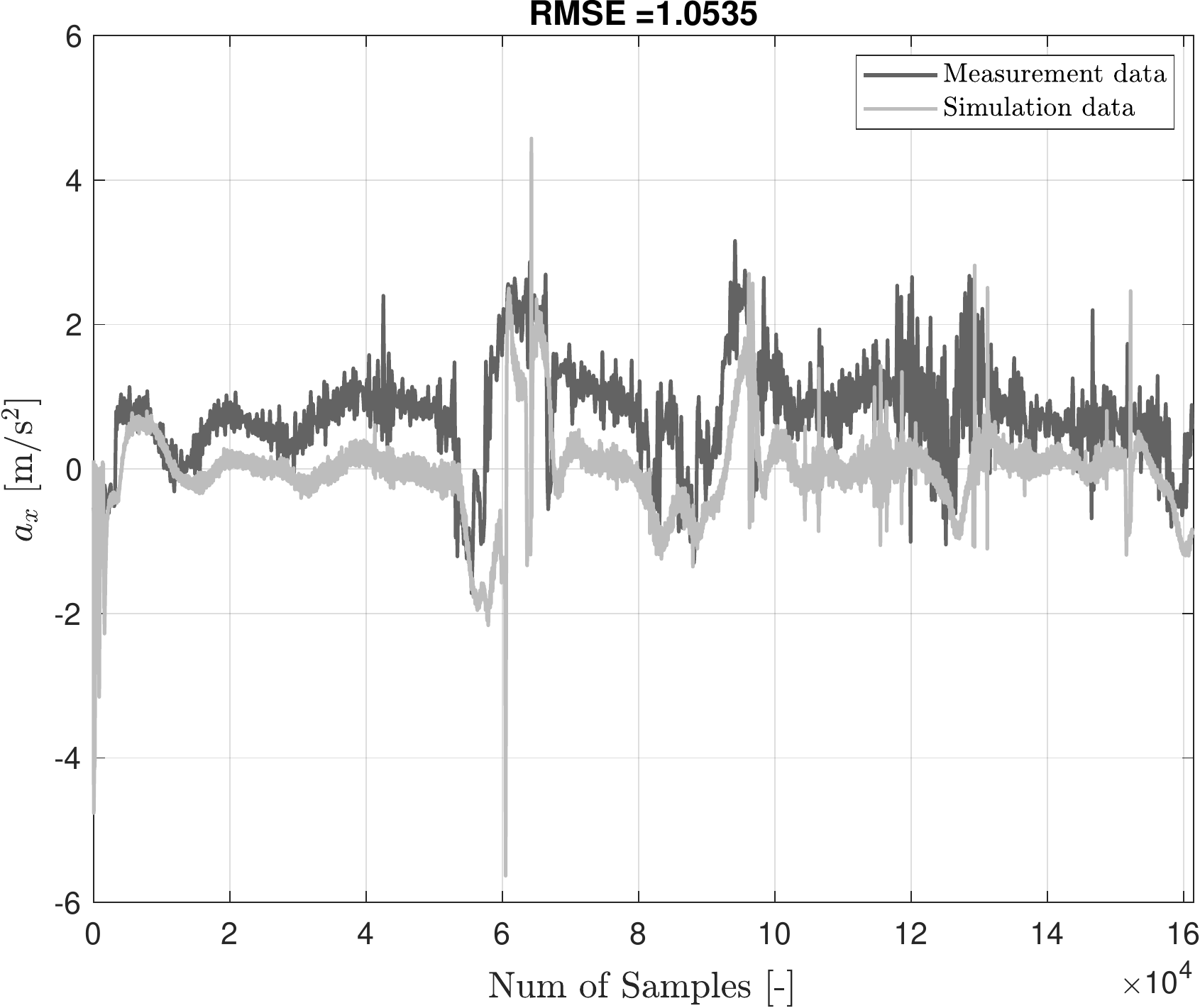}}
    \hfill
  \subfloat[]{%
       \includegraphics[width=0.33\linewidth]{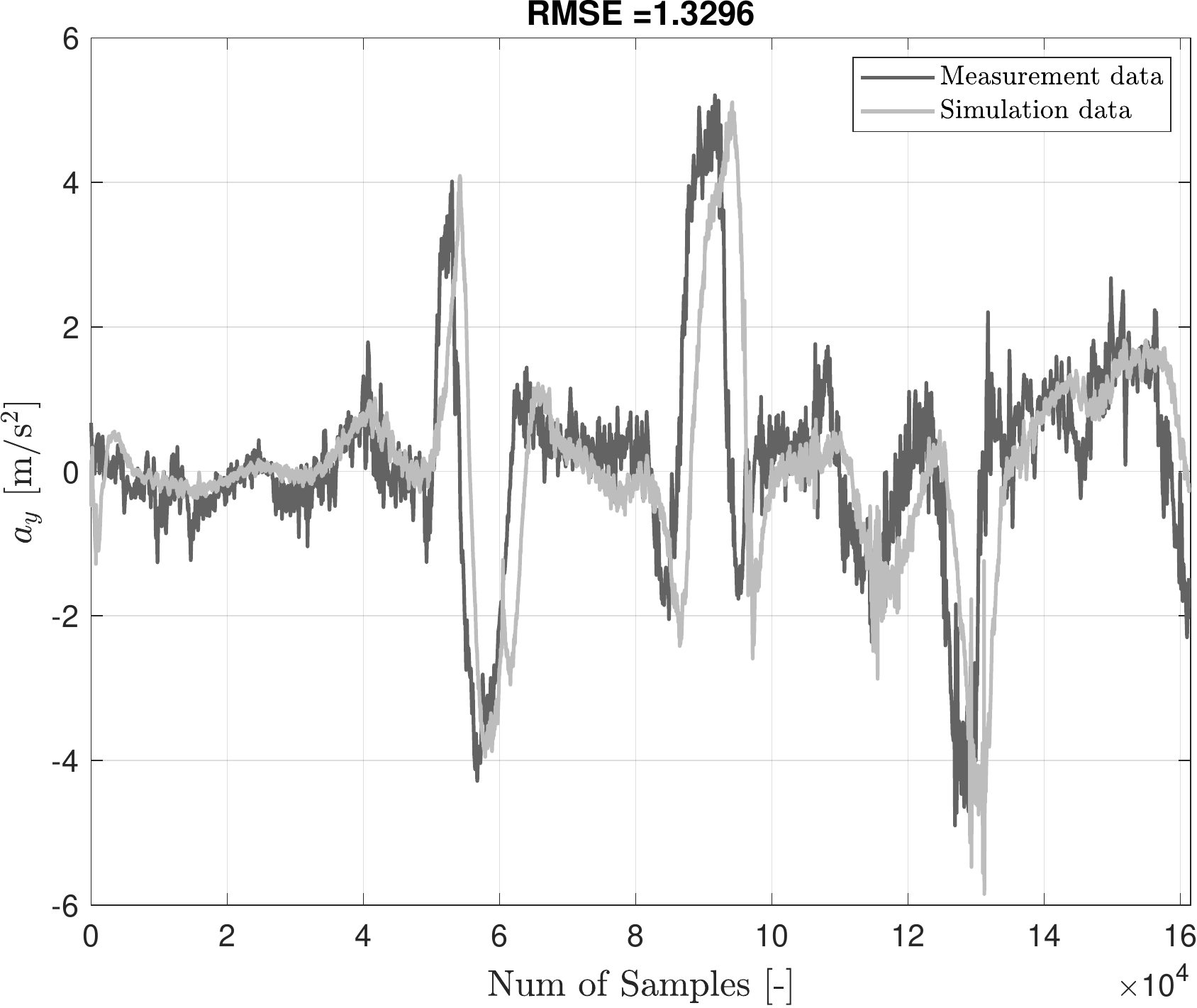}}
  \hfill
  \subfloat[]{%
        \includegraphics[width=0.33\linewidth]{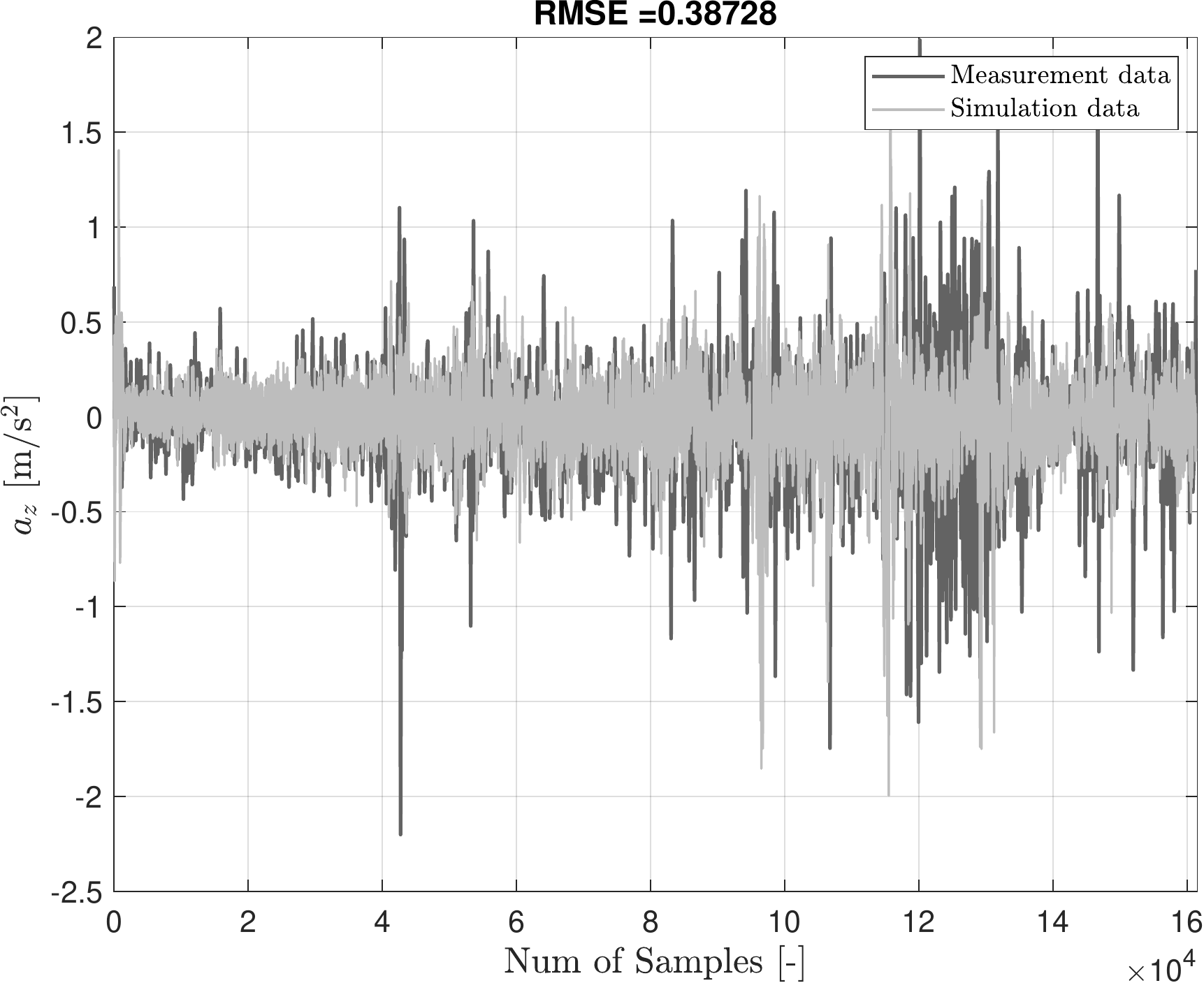}}
    \\
  \subfloat[]{%
        \includegraphics[width=0.33\linewidth]{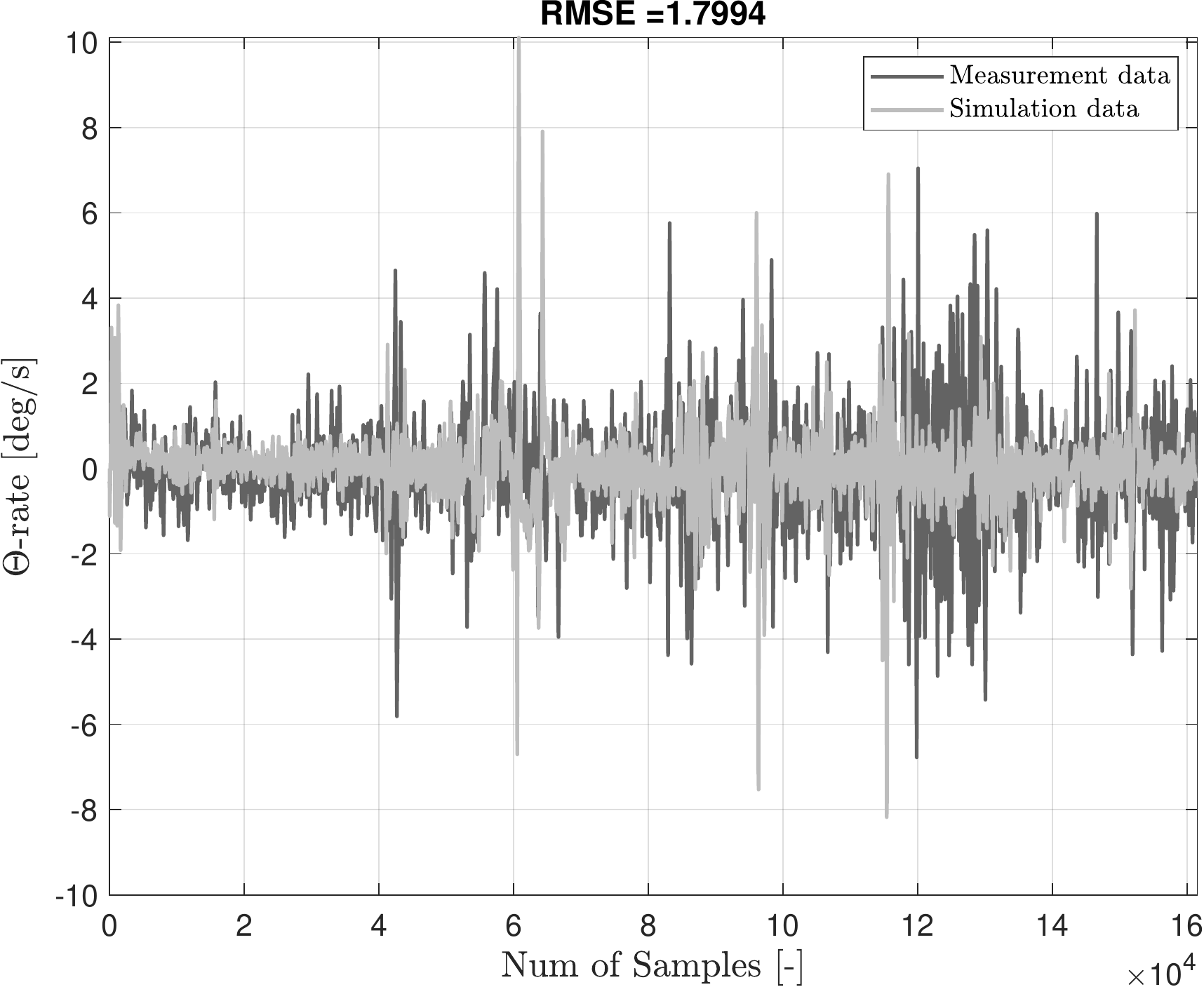}}
  \hfill
  \subfloat[]{%
        \includegraphics[width=0.33\linewidth]{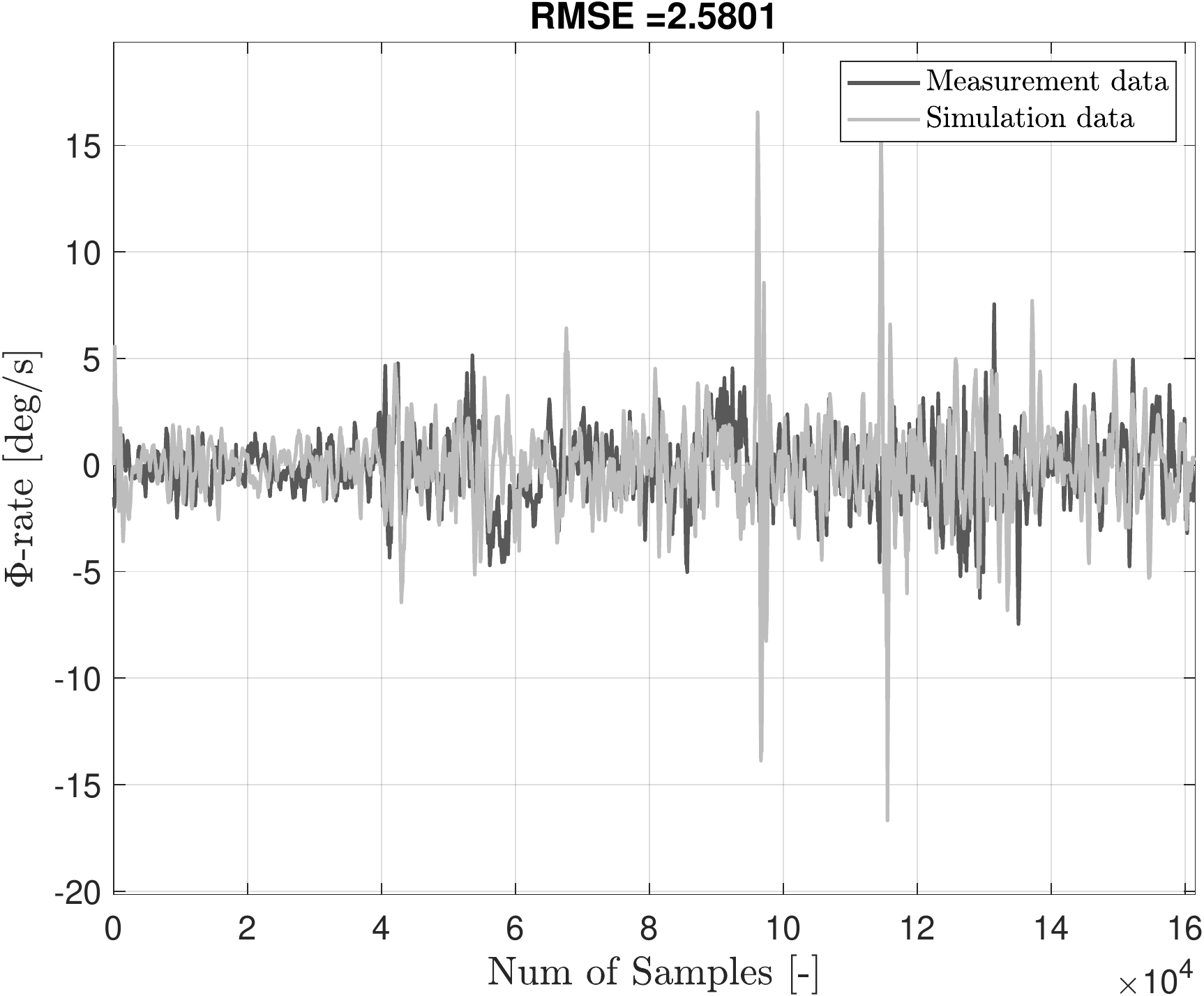}}
    \hfill
  \subfloat[]{%
        \includegraphics[width=0.33\linewidth]{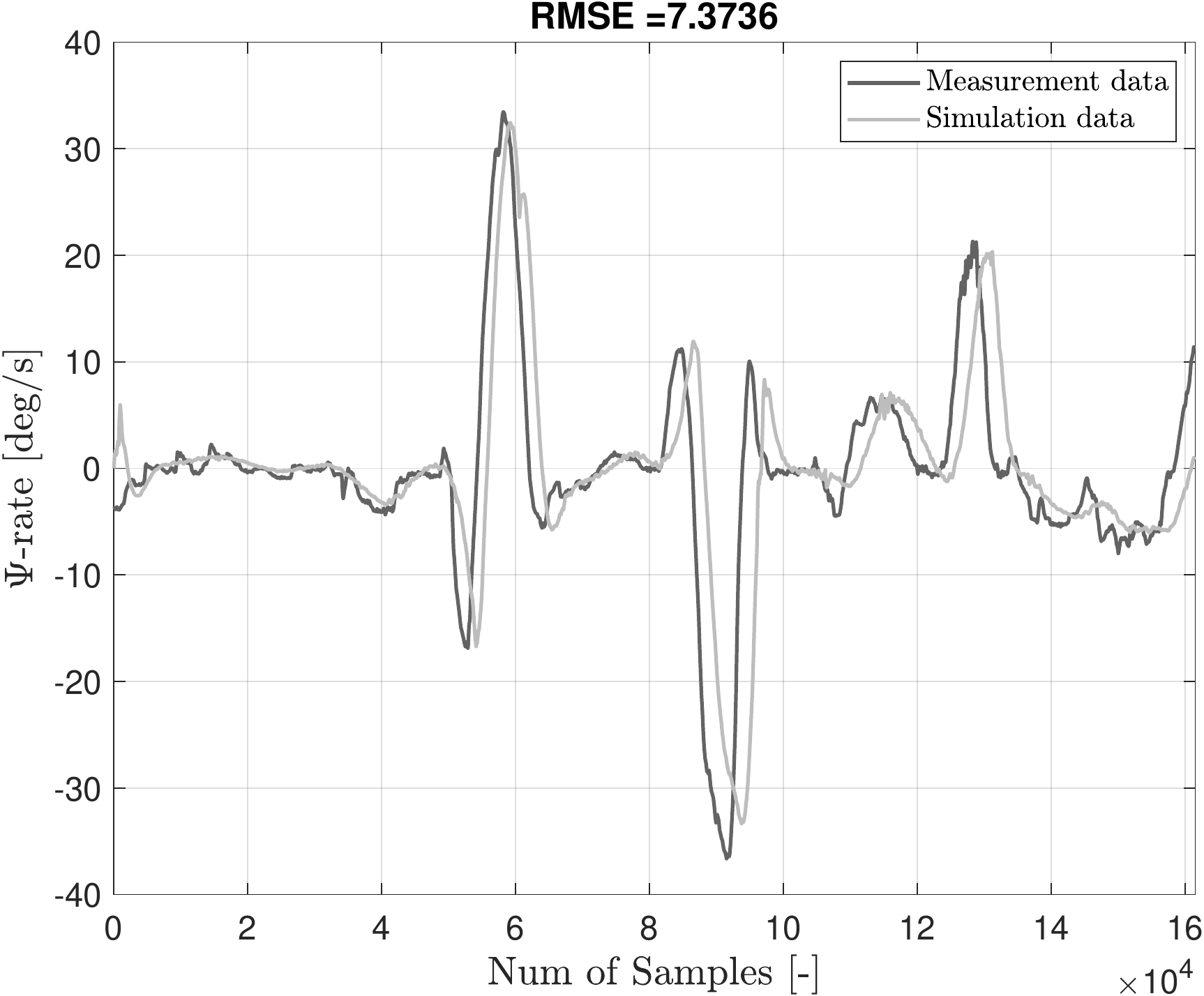}}
  \caption{Comparison of real-world measurements and non-optimized simulation outputs for $Ts{\rm 3}$: (a) $a_x$ (RMSE = 1.0535); (b) $a_y$ (RMSE = 1.3296); (c) $a_z$ (RMSE = 0.3873); (d) $\Theta$-rate (RMSE = 1.7994); (e) $\Phi$-rate (RMSE = 2.5801); (f) $\Psi$-rate (RMSE = 7.3736).}
  \label{fig:opti_plots} 
\end{figure}

In order to validate the various model implementations of the simulator, a comparison between the measurement data and simulation outputs is performed. Table~\ref{tab:baseline_res} provides the NRMSE for each test site. By averaging the NRMSE values per site, $Ts{\rm 1}$ demonstrates the highest average error with $0.1455$, followed by $0.1305$ and $0.1198$ for $Ts{\rm 2}$ and $Ts{\rm 3}$, respectively. These values justify the decision to optimize the models on $Ts{\rm 1}$. Furthermore, Fig.~\ref{fig:opti_plots} depicts a comparison of such corresponding signals from  $Ts{\rm 3}$. As the NRMSE allows comparability of the different errors and hinders the derivation of the error magnitude for one quantity, we state the RMSE in Fig.~\ref{fig:opti_plots}. The comparison plots show that the simulator models real-world behavior with sufficient accuracy and that there is room for improvement by performing an optimization procedure for model parameters.
\begin{table}[!b]
	\centering
	\small
	\caption{Comparison of real-world data and non-optimized model outputs.}
	\begin{tabular}{lrrrrrrr}
		\toprule
			& \multicolumn{1}{l}{$\mathrm{NRMSE}_{a_x}$} & \multicolumn{1}{l}{$\mathrm{NRMSE}_{a_y}$} & \multicolumn{1}{l}{$\mathrm{NRMSE}_{a_z}$} & \multicolumn{1}{l}{$\mathrm{NRMSE}_\Phi$} & \multicolumn{1}{l}{$\mathrm{NRMSE}_\Theta$} & \multicolumn{1}{l}{$\mathrm{NRMSE}_\Psi$} \\
			\midrule
		Ts1 & 0.1145 & 0.0960 & 0.1217 & 0.3055 & 0.1368 & 0.0982\\
		Ts2 & 0.1222 & 0.1028 & 0.0939 & 0.2726 & 0.1006 & 0.0906\\
		Ts3 & 0.2070 & 0.0851 & 0.0909 & 0.1564 & 0.1207 & 0.0588\\
		\bottomrule
	\end{tabular}%
	\label{tab:baseline_res}%
\end{table}%

\begin{table}[!b]
	\centering
	\small
	\caption{Optimization parameters with lower and upper constraints values.}
	\begin{tabular}{llrr}
		\toprule
		Variable & Unit  & \multicolumn{1}{l}{Lower Bound} & \multicolumn{1}{l}{Upper Bound} \\
		\midrule
		Spring stiffness front $K_{\mathrm{sf}}$ & [N/m] & 15000 & 40000 \\
		Spring stiffness rear $K_{\mathrm{sr}}$& [N/m] & 15000 & 40000 \\
		Tire friction value $\mu_{\mathrm{T}}$ & [-]   & 0.7   & 1.4 \\
		Tire radial stiffness $K_{\mathrm{Tr}}$ & [N/m] & 250000 & 400000 \\
		Tire radial damping $d_{\mathrm{Tr}}$& [Ns/m] & 4000  & 7000 \\
		\bottomrule
	\end{tabular}%
	\label{tab:constraints}%
\end{table}%

\begin{table}[!b]
	\centering
	\small
	\caption{Comparison of non-optimized and optimized models -- $Ts{\rm 1}$.}
	\begin{tabular}{lrrrrrr}
		\toprule
			& \multicolumn{1}{l}{$\mathrm{NRMSE}_{a_x}$} & \multicolumn{1}{l}{$\mathrm{NRMSE}_{a_y}$} & \multicolumn{1}{l}{$\mathrm{NRMSE}_{a_z}$} & \multicolumn{1}{l}{$\mathrm{NRMSE}_\Phi$} & \multicolumn{1}{l}{$\mathrm{NRMSE}_\Theta$} & \multicolumn{1}{l}{$\mathrm{NRMSE}_\Psi$} \\
			\midrule
		Ref & 0.1145 & 0.0960 & 0.1217 & 0.3055 & 0.1368 & 0.0982 \\
		Opt & 0.1131 & 0.0951 & 0.1099 & 0.3082 & 0.1301 & 0.0977 \\
		\midrule
		$\pm$ [\%] & 1.2226 & 1.0168 & 9.7166 & -0.8606 & 4.8976 & 0.5086 \\
		\bottomrule
	\end{tabular}%
	\label{tab:opti_res_ts1}%
\end{table}%

Given the problem defined in equation~(\ref{eq:nonlinlsq}), measurement data are defined as the reference $y(x)$. The output of the simulation as a function of $x$ and $p$ is defined as $f(x,p)$. The residuals shown in Table~\ref{tab:baseline_res} represent the corresponding $r(x,p)$. We utilize the defined optimization problem in~(\ref{eq:opti}) and constrain the problem with reasonable lower and upper bound values presented in Table~\ref{tab:constraints}. 

In the following, the optimization procedure is applied to test site $Ts{\rm 1}$. A simulation with standard model settings is defined as a benchmark (reference) case referred to as \text{Ref}. The optimized results are indicated with the abbreviation \text{Opt}. The results for $Ts{\rm 1}$ are depicted in Table~\ref{tab:opti_res_ts1}. 

All three directions of accelerations $a_x$, $a_y$, and $a_z$ achieve an improvement of $1.22\%$, $1.02 \%$, and $9.72\%$, respectively. Note that acceleration $a_z$, is of particular interest here, as it induces forces that mainly originate at the road surface and then transferred to the human body.
To further justify the optimization concept, the tuned models obtained from $Ts{\rm 1}$ are then tested on sites $Ts{\rm 2}$ and $Ts{\rm 3}$. Tables~\ref{tab:opti_res_ts2} and~\ref{tab:opti_res_ts3} present the corresponding results. Once again, the three error values of the acceleration parameters have decreased. The improvement of $5.79\%$ and $3.40\%$ in the $z$-direction demonstrates the effectiveness of the optimization procedure. The negative gradient of improvement from $Ts{\rm 1}$ to $Ts{\rm 3}$ can be justified by the different sites' characteristics. Note that by optimizing each test site separately, these values could be further improved. 

\begin{table}[b]
	\centering
	\small
	\caption{Comparison of non-optimized and optimized models -- $Ts{\rm 2}$.}
	\begin{tabular}{lrrrrrr}
		\toprule
			& \multicolumn{1}{l}{$\mathrm{NRMSE}_{a_x}$} & \multicolumn{1}{l}{$\mathrm{NRMSE}_{a_y}$} & \multicolumn{1}{l}{$\mathrm{NRMSE}_{a_z}$} & \multicolumn{1}{l}{$\mathrm{NRMSE}_\Phi$} & \multicolumn{1}{l}{$\mathrm{NRMSE}_\Theta$} & \multicolumn{1}{l}{$\mathrm{NRMSE}_\Psi$} \\
			\midrule
		Ref & 0.1222 & 0.1028 & 0.0939 & 0.2726 & 0.1006 & 0.0906 \\
		Opt & 0.1211 & 0.1020 & 0.0885 & 0.2530 & 0.0972 & 0.0902 \\
		\midrule
		$\pm$ [\%] & 0.9240 & 0.7972 & 5.7935 & 7.1745 & 3.3836 & 0.4205 \\
		\bottomrule
	\end{tabular}%
	\label{tab:opti_res_ts2}%
\end{table}%

\begin{table}[t]
	\centering
	\small
	\caption{Comparison of non-optimized and optimized models -- $Ts{\rm 3}$.}
	\begin{tabular}{lrrrrrr}
		\toprule
			& \multicolumn{1}{l}{$\mathrm{NRMSE}_{a_x}$} & \multicolumn{1}{l}{$\mathrm{NRMSE}_{a_y}$} & \multicolumn{1}{l}{$\mathrm{NRMSE}_{a_z}$} & \multicolumn{1}{l}{$\mathrm{NRMSE}_\Phi$} & \multicolumn{1}{l}{$\mathrm{NRMSE}_\Theta$} & \multicolumn{1}{l}{$\mathrm{NRMSE}_\Psi$} \\
			\midrule
		Ref & 0.2070 & 0.0851 & 0.0909 & 0.1564 & 0.1207 & 0.0588 \\
		Opt & 0.2063 & 0.0845 & 0.0878 & 0.1498 & 0.1167 & 0.0584 \\
		\midrule
		$\pm$ [\%] & 0.3573 & 0.6356 & 3.3981 & 4.2375 & 3.3140 & 0.7390 \\
		\bottomrule
	\end{tabular}%
	\label{tab:opti_res_ts3}%
\end{table}%

\subsection{Stochastic simulation set-up}
\label{sec:stochastic_sim_set_up}
The speed and driving trajectory of human drivers frequently deviate from ideal trajectories; also, weather conditions change regularly in reality; all these factors of influence are modeled in our simulations. The lateral lane position of a vehicle $l_p$, speed profile deviation $v_{\rm dev}$, and friction coefficient of road surface $\mu_{\rm rs}$ vary for each simulation run. The corresponding PDFs with the associated parameters are provided in Table~\ref{tab:lhs_input}. 

\begin{table}[!b]
  \centering
  \caption{Representation of the probabilistic input parameters. The variables $l$ and $c$ denote the length and curvature of the test site, respectively.}
    \begin{tabular}{lllll}
    \toprule
    \multicolumn{1}{l}{Test Site} & Variable & Distribution & \multicolumn{2}{c}{Parameters} \\
    \midrule
    \multicolumn{1}{l}{$Ts{\rm 1}$} & $v_{\rm dev}$  & Gaussian & $\mu=0$  & $\sigma=0.2$ \\
         $l=3.0$km & $l_p$ & Gaussian & $\mu=0$   & $\sigma=0.2$ \\
         $c$ = low & $\mu_{\rm rs}$  & Uniform & $a=0.6$ & $b=1.0$ \\
    \midrule
    \multicolumn{1}{l}{$Ts{\rm 2}$} & $v_{\rm dev}$  & Gaussian & $\mu=0$  & $\sigma=0.2$ \\
         $l = 1.7$km & $l_p$ & Gaussian &  $\mu=0$   & $\sigma=0.2$ \\
         $c$ = medium & $\mu_{\rm rs}$  & Uniform & $a=0.6$ & $b=1.0$ \\
    \midrule
    \multicolumn{1}{l}{$Ts{\rm 3}$} & $v_{\rm dev}$  & Gaussian & $\mu=-5$ & $\sigma=0.2$ \\
          $l = 2.0$km & $l_p$ & Gaussian & $\mu=0$   &$\sigma=0.2$ \\
          $c$ = high & $\mu_{\rm rs}$  & Uniform & $a=0.6$ & $b=1.0$ \\
    \bottomrule
    \end{tabular}%
  \label{tab:lhs_input}%
\end{table}%

It should be pointed out that the normal distribution for the speed deviation of $Ts{\rm 3}$ is shifted, with parameters $\mathcal{N}(-5, 0.2)$, resulting mostly in negative speed deviation samples. This modeling approach has been chosen due to the high curvature of the specific test site. Note that positive speed deviations would have led to a high number of vehicles running off the road during simulations. To this end, 4.000 input samples have been created for every test site, and the corresponding simulations have been performed. This results in vehicle dynamics output data from 12.000 simulations in total.  

\subsection{Ride comfort results}
\label{sec:ride_comfort_results}
As proposed in Section~\ref{sec:ride_comfort}, the three ride comfort determination strategies are applied to the simulation output (i.e., the simulated vehicle dynamics data) and the IRI measurement data. Afterward, we compare the thresholding and the ISO 2631 strategy against the IRI comfort classifications. 

\begin{figure}[htpb]
    \centering
  \subfloat[]{%
       \includegraphics[width=0.5\linewidth]{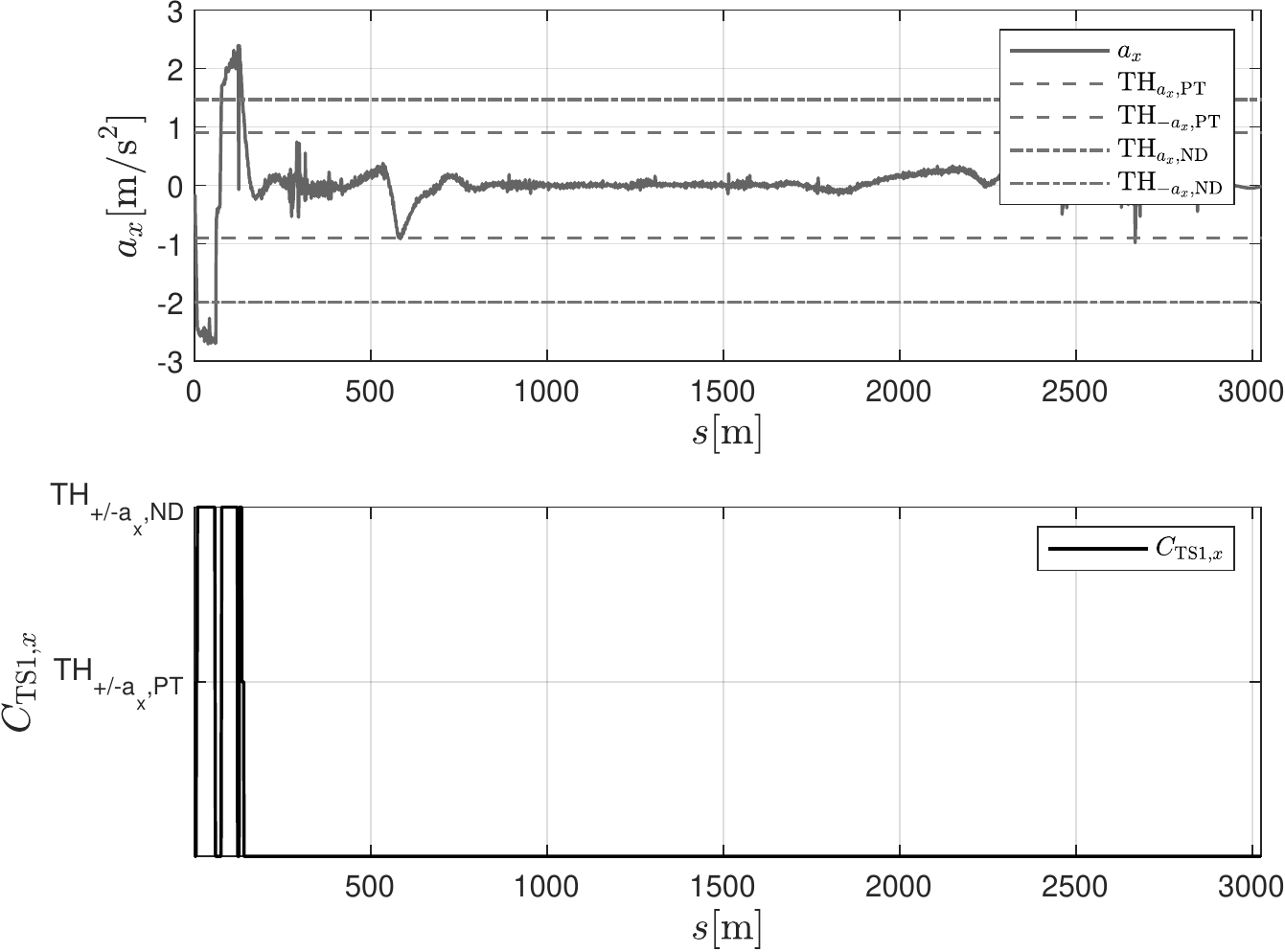}}
    \hfill
  \subfloat[]{%
       \includegraphics[width=0.5\linewidth]{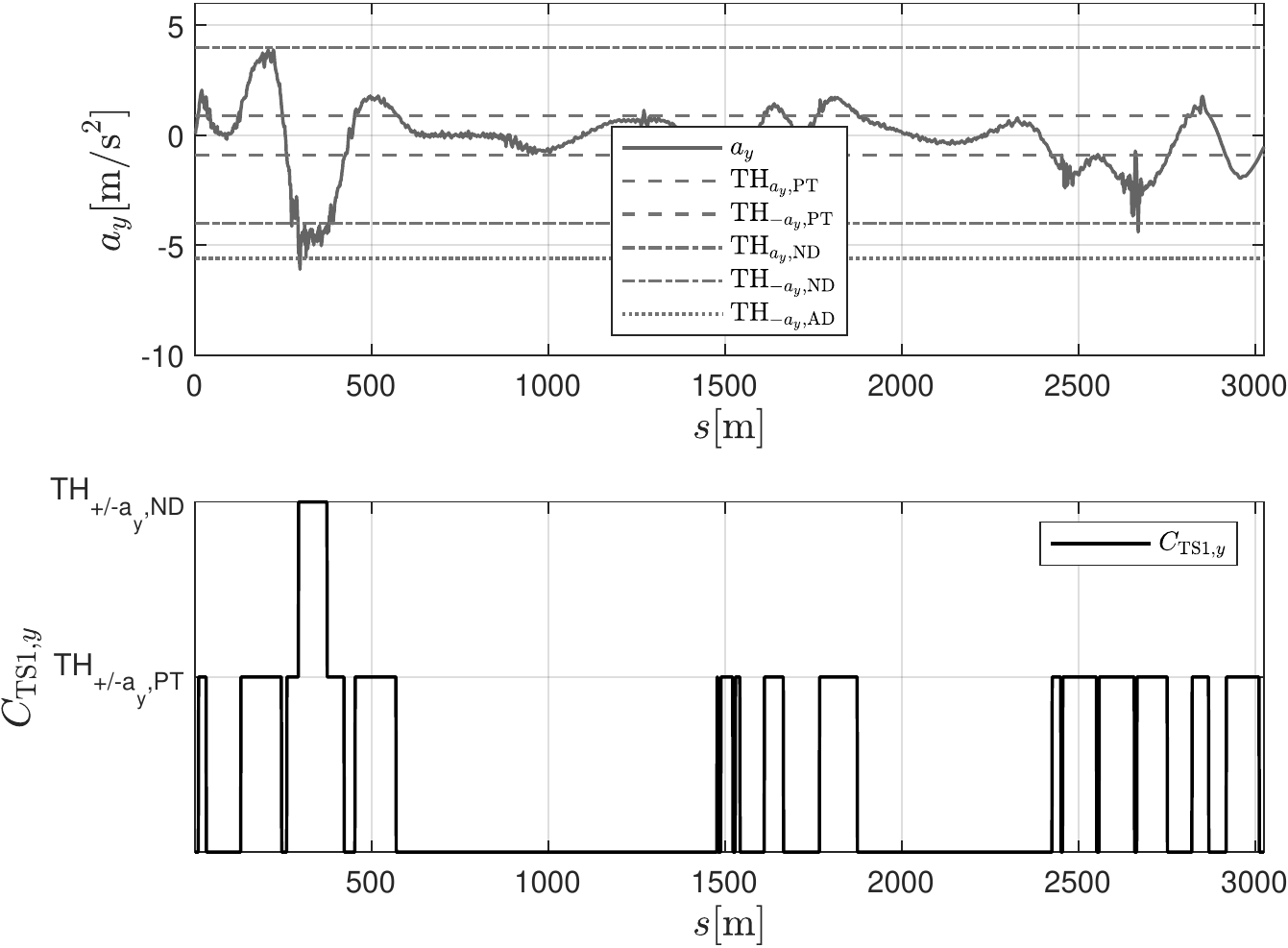}}
   \\
   \subfloat[]{%
       \includegraphics[width=0.5\linewidth]{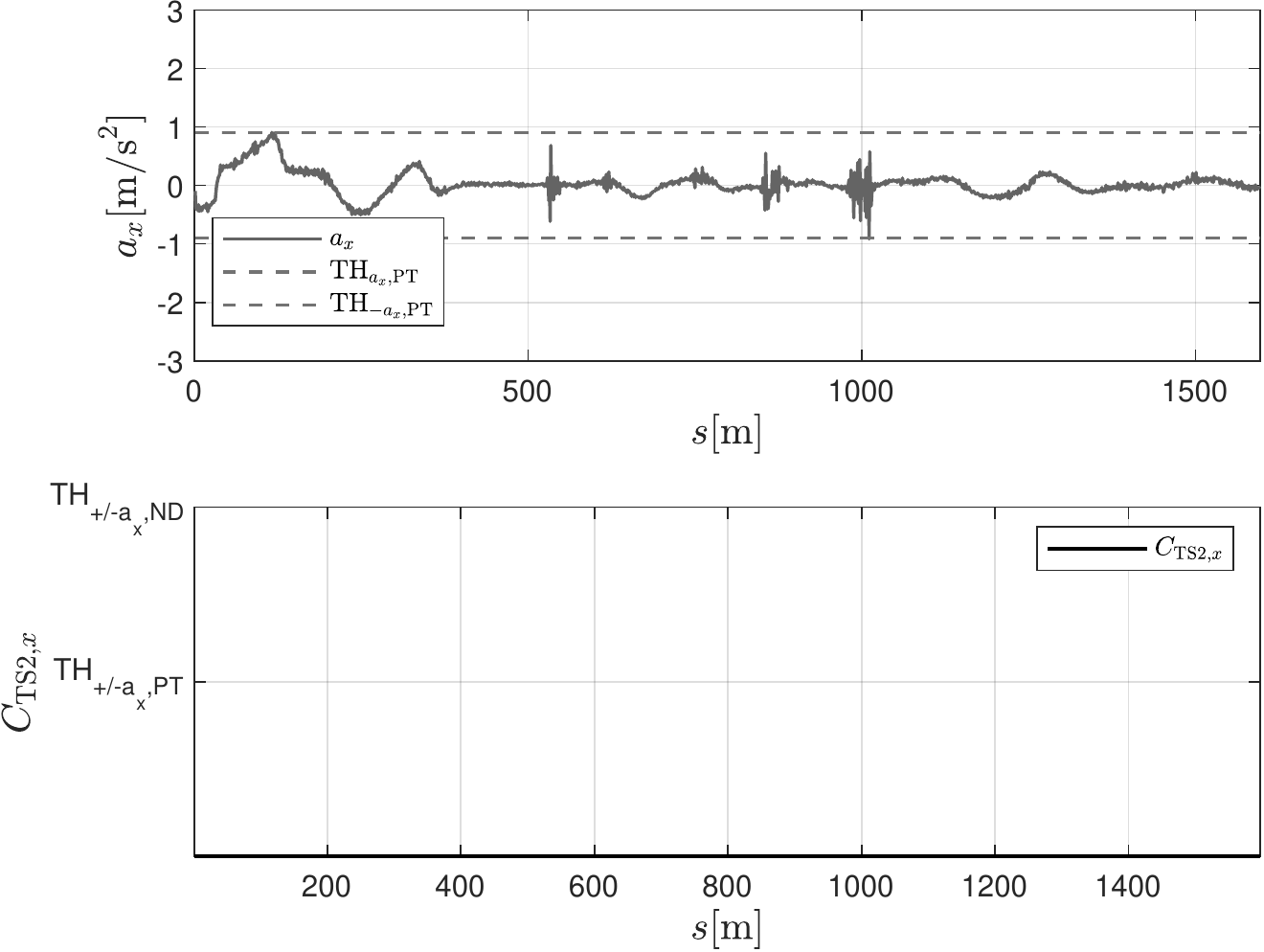}}
    \hfill
  \subfloat[]{%
       \includegraphics[width=0.5\linewidth]{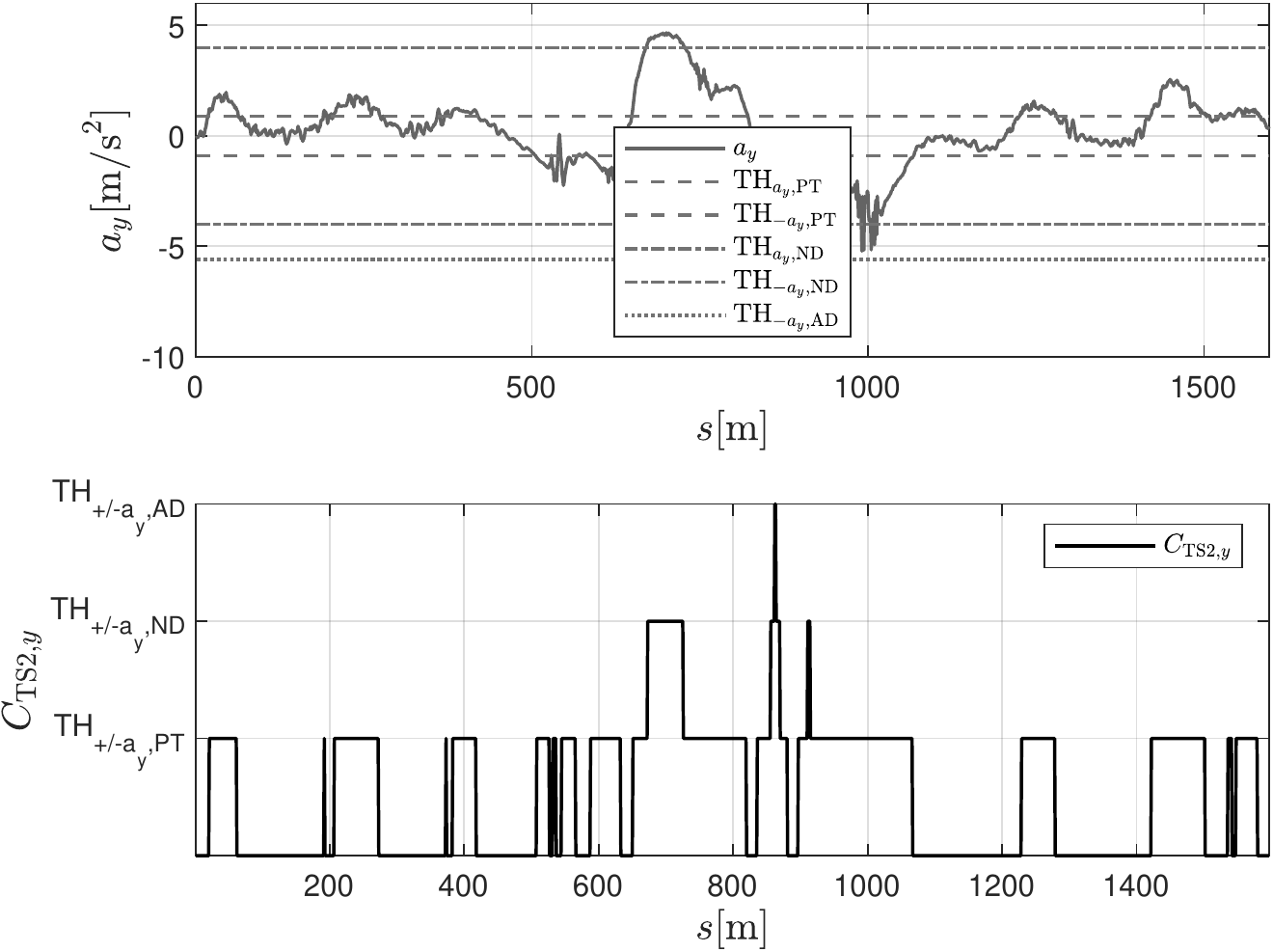}}
   \\
   \subfloat[]{%
       \includegraphics[width=0.5\linewidth]{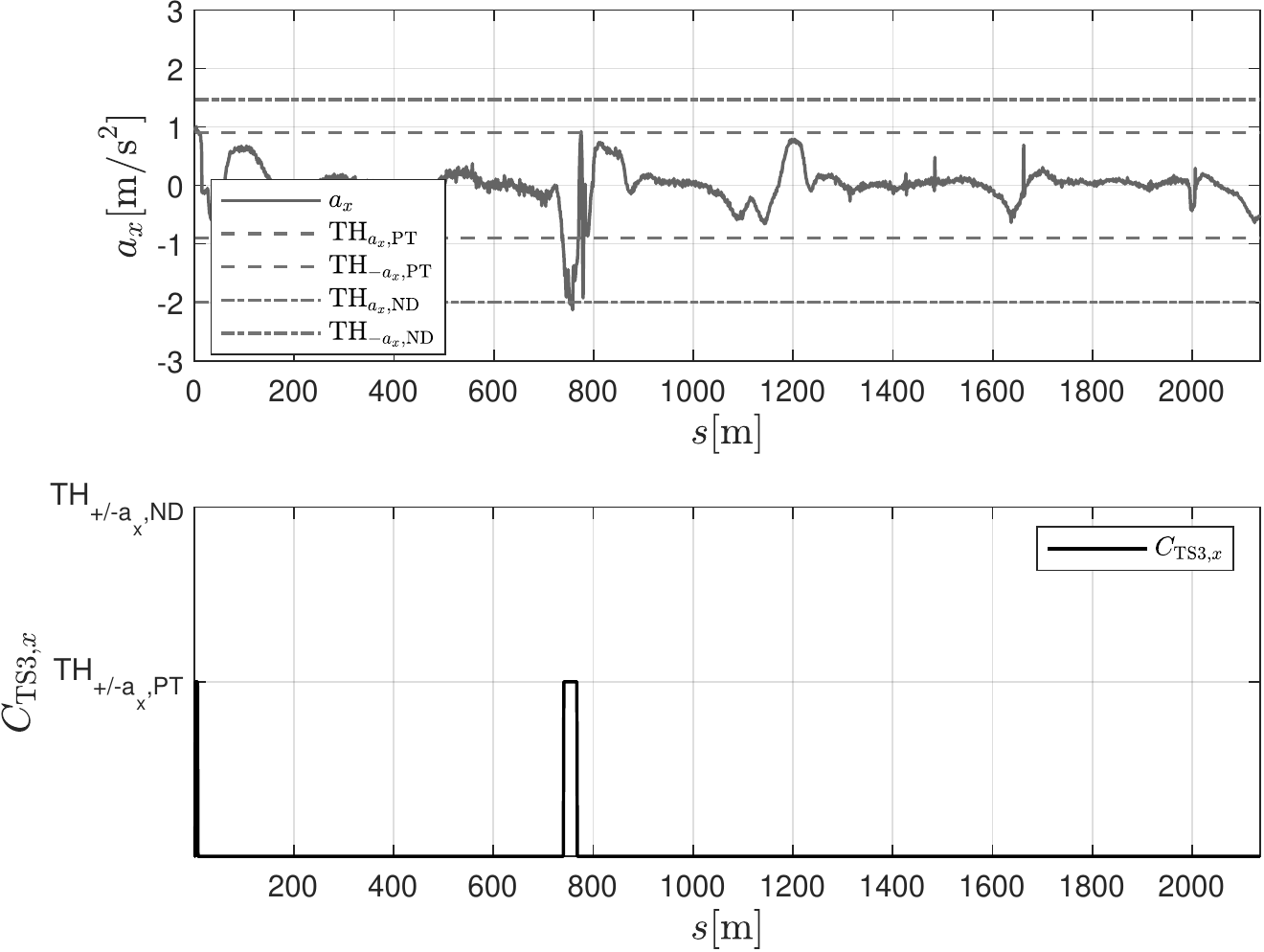}}
    \hfill
  \subfloat[]{%
       \includegraphics[width=0.5\linewidth]{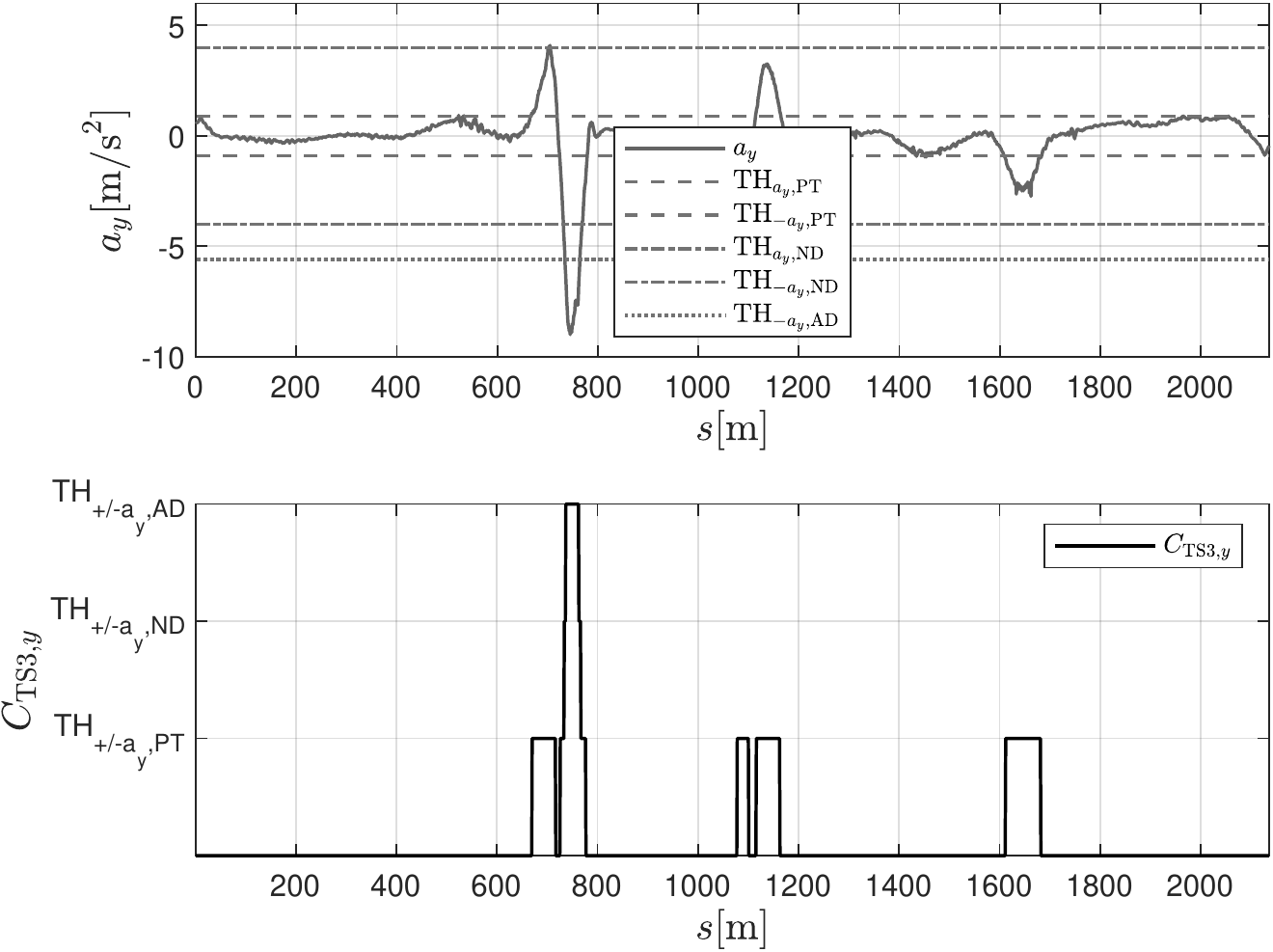}}
   \\
  \caption{Thresholding results for the accelerations $a_x$ and $a_y$ of  $Ts{\rm 1}$,  $Ts{\rm 2}$, and  $Ts{\rm 3}$.}
  \label{fig:thresholding_results_1} 
\end{figure}

\begin{figure}[htpb]
    \centering
    \subfloat[]{%
        \includegraphics[width=0.5\linewidth]{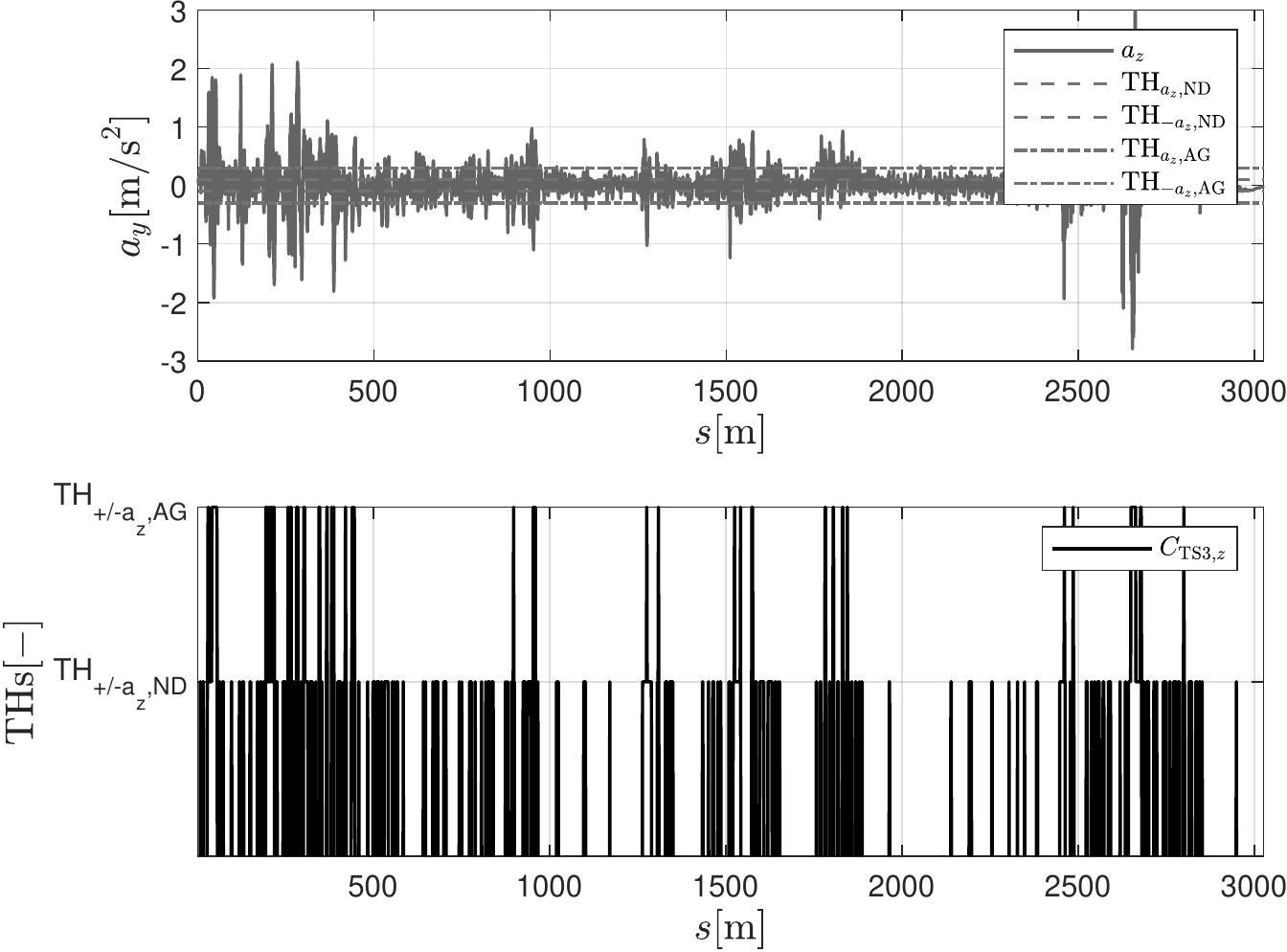}}
    \\

    \subfloat[]{%
        \includegraphics[width=0.5\linewidth]{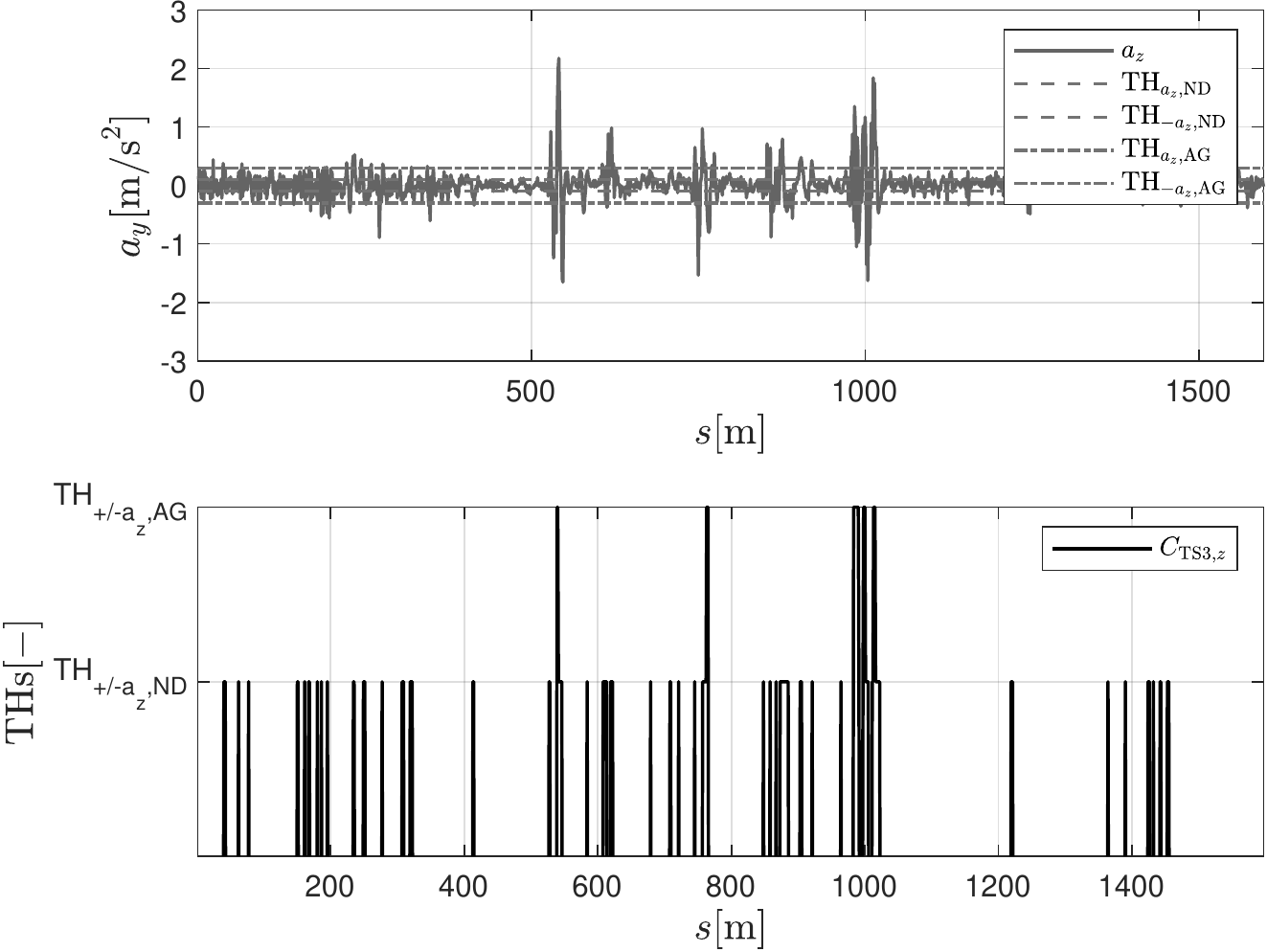}}
    \\
  
    \subfloat[]{%
        \includegraphics[width=0.5\linewidth]{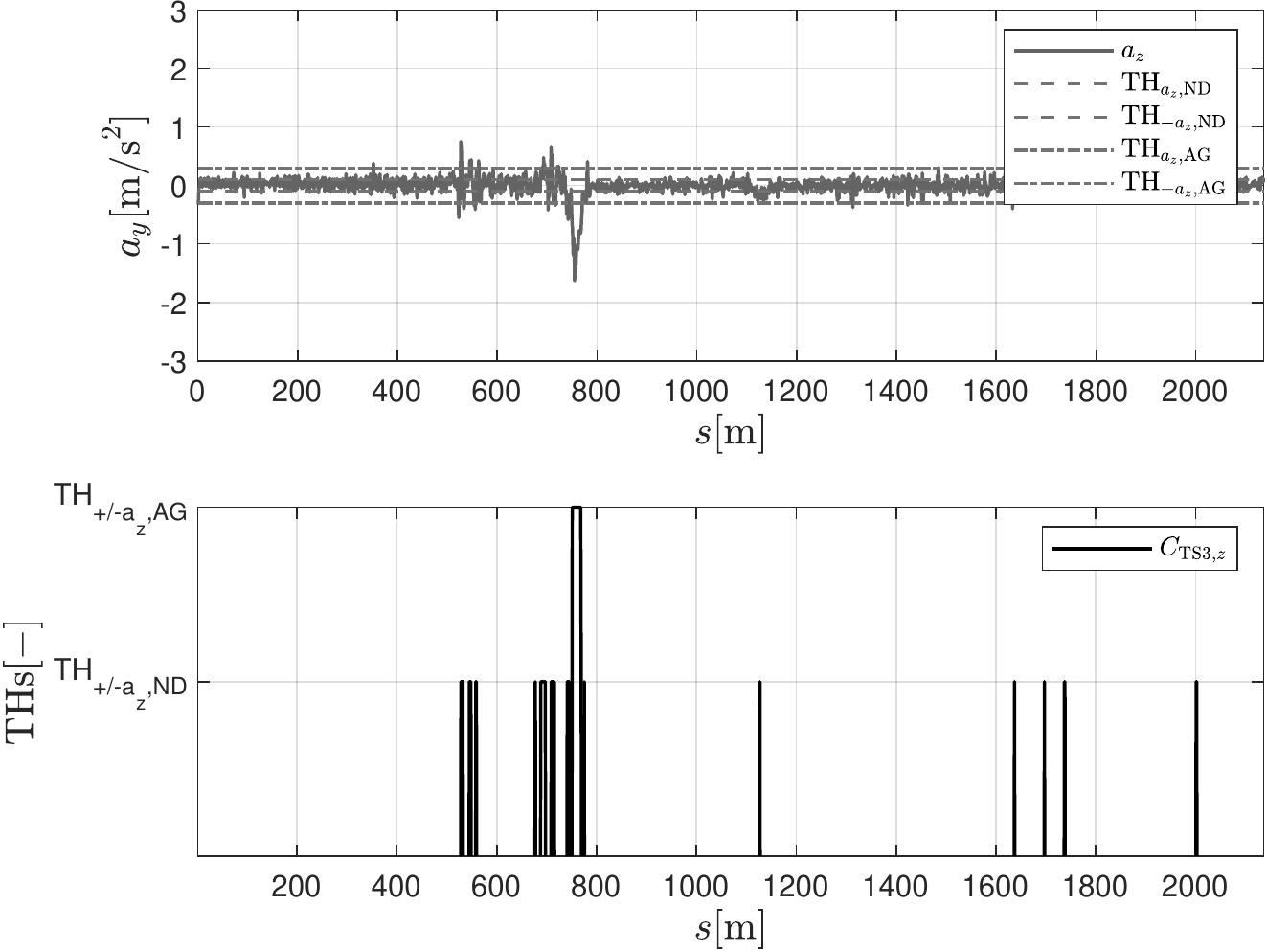}}
    \\   
   \caption{Thresholding results for the accelerations $a_z$ of  $Ts{\rm 1}$,  $Ts{\rm 2}$, and  $Ts{\rm 3}$..}
  \label{fig:thresholding_results_2} 
\end{figure}

First, we apply the thresholding procedure from Section~\ref{sec:thresholding}. Before applying the methodology, the data needs to be pre-processed; i.e., signals must be transformed from time to space domain. Afterward, the thresholds are applied to the acceleration outputs $a_x$, $a_y$, and $a_z$ of all three test sections. Figures~\ref{fig:thresholding_results_1} and~\ref{fig:thresholding_results_2} depict the acceleration signals, the corresponding thresholds (only those thresholds which are applicable are utilized) and the resulting classification signal $C_{\mathrm{Ts}_i,\{x,y,z\}}$, where $\{x,y,z\}$ stands for the set of investigated acceleration $a_x$, $a_y$, and $a_z$. 

A qualitative inspection of the classification outputs $C_{\mathrm{Ts}_i,\{x\}}$ shows that the aggregated accelerations in the $x$-direction do not present significant deviations. Only two signal peaks at the beginning of $\mathrm{TS}_1$ exceeding the threshold of normal driving and two signal peaks at the beginning and around 750 meters on $\mathrm{TS}_3$ exceeding the threshold for public transportation comfort are identified. The $a_x$ signal of $\mathrm{TS}_2$ does not exceed any threshold, and hence no criticalities are found. As the simulations are performed without any traffic interactions, and consequently the vehicle can move in free-flow conditions, the number of signal peaks of $C_{\mathrm{Ts}_i,\{x\}}$ are expected. Note that fluctuations in the acceleration signals do occur as the road surface does influence the ACC, trying to maintain a velocity as close as possible to the desired one. The quantitative classification results draw the same picture and are collected in Table~\ref{tab:thresholding_table}. Every test site is processed sequentially for critical sections, with a search window defined by an average car length of $l_{\rm cr} = 5$m. This value has been assumed to be reasonable in order to produce negative ride experiences; nevertheless, this parameter can be changed in the generic framework. The variable $C$ represents the detected critical road sections; $R_{\mathrm{c}}$ denotes the critical ratio of the test section in \%; $N$ and  $R_{\mathrm{n}}$ are the equivalent variables for the non-critical detections. Results in the $x$-direction are shown for $Ts{\rm 1}$, $Ts{\rm 2}$ and $Ts{\rm 3}$ and for $a_x$, $a_y$ and $a_z$, respectively. Along with the qualitative inspection, $Ts{\rm 1}$ shows 24 critical sections, which correspond to $3.97\%$ of the test section's length that exceed the threshold for public transportation. $Ts{\rm 3}$ shows 7 critical sections corresponding to $1.64\%$ of its length. 

\begin{table}[!b]
  \centering
  
  \caption{Detected critical ride comfort road sections based on the thresholding method for all test sites with a critical length $l_{\rm cr} = 5$m. Note that for $a_z \rightarrow \pm a_{x,\mathrm{ND}} = \pm a_{x,\mathrm{PT}}$. Note that PT = Public Transportation, ND = Normal driving, AG = Aggressive driving. $C$ and $N$ represent the critical and non-critical road sections; $R_{\mathrm{c}}$ and $R_{\mathrm{n}}$ the corresponding relative quantities.}
  \begin{adjustbox}{width=\columnwidth,center}
    \begin{tabular}{l|rrrr|rrrr|rrrr}
    \toprule 
         \multicolumn{1}{l}{}  & \multicolumn{12}{c}{Longitudinal acceleration $a_x$} \\
          
         \multicolumn{1}{l}{} & \multicolumn{4}{c}{$Ts{\rm 1}$}       & \multicolumn{4}{c}{$Ts{\rm 2}$}       & \multicolumn{4}{c}{$Ts{\rm 3}$} \\
          
          \multicolumn{1}{l}{} & \multicolumn{1}{l}{$C$ [-]} & \multicolumn{1}{l}{$R_{\mathrm{c}}$ [\%]} & \multicolumn{1}{l}{$N$ [-]} & \multicolumn{1}{l}{$R_{\mathrm{n}}$ [\%]} & \multicolumn{1}{l}{$C$ [-]} & \multicolumn{1}{l}{$R_{\mathrm{c}}$ [\%]} & \multicolumn{1}{l}{$N$ [-]} & \multicolumn{1}{l}{$R_{\mathrm{n}}$ [\%]} & \multicolumn{1}{l}{$C$ [-]} & \multicolumn{1}{l}{$R_{\mathrm{c}}$ [\%]} & \multicolumn{1}{l}{$N$ [-]} & \multicolumn{1}{l}{$R_{\mathrm{n}}$ [\%]} \\
          \midrule
            $\pm a_{x,\mathrm{PT}}$ & \textbf{24.00}  &  \textbf{3.97} & 581.00 &  96.03    &     0.00    &     0.00 & 319.00 & 100.00  &  7.00  &  1.64 & 420.00 &  98.36 \\
            $\pm a_{x,\mathrm{ND}}$ & \textbf{21.00}  &  \textbf{3.47} & 584.00 &  96.53    &     0.00    &     0.00 & 319.00 & 100.00  &     0.00  &      0.00 & 427.00 & 100.00 \\
            $\pm a_{x,\mathrm{AG}}$ &    0.00  &     0.00 &  605.00 & 100.00    &     0.00    &     0.00 & 319.00 & 100.00  &     0.00  &      0.00 & 427.00 & 100.00 \\

    \midrule
          \multicolumn{1}{l}{} & \multicolumn{12}{c}{Lateral acceleration $a_y$} \\
          
           \multicolumn{1}{l}{}& \multicolumn{4}{c}{$Ts{\rm 1}$}       & \multicolumn{4}{c}{$Ts{\rm 2}$}       & \multicolumn{4}{c}{$Ts{\rm 3}$} \\
          
          \multicolumn{1}{l}{} & \multicolumn{1}{l}{$C$ [-]} & \multicolumn{1}{l}{$R_{\mathrm{c}}$ [\%]} & \multicolumn{1}{l}{$N$ [-]} & \multicolumn{1}{l}{$R_{\mathrm{n}}$ [\%]} & \multicolumn{1}{l}{$C$ [-]} & \multicolumn{1}{l}{$R_{\mathrm{c}}$ [\%]} & \multicolumn{1}{l}{$N$ [-]} & \multicolumn{1}{l}{$R_{\mathrm{n}}$ [\%]} & \multicolumn{1}{l}{$C$ [-]} & \multicolumn{1}{l}{$R_{\mathrm{c}}$ [\%]} & \multicolumn{1}{l}{$N$ [-]} & \multicolumn{1}{l}{$R_{\mathrm{n}}$ [\%]} \\
          \midrule
 $\pm a_{y,\mathrm{PT}}$ &  221.00  &  36.53 &  384.00 &   63.47 &  \textbf{163.00}  &  \textbf{51.10} &  156.00 &   48.90  &  50.00  &  11.71 &  377.00  &  88.29 \\
$\pm a_{y,\mathrm{ND}}$ &   17.00  &   2.81 &  588.00 &   97.19 &   \textbf{15.00}  &   \textbf{4.70} &  304.00  &  95.30  &   6.00  &   1.41 &  421.00  &  98.59 \\
$\pm a_{y,\mathrm{AG}}$ &    0.00  &        0.00 &   605.00 &  100.00  &   1.00  &   0.31 &  318.00 &   99.69  &   \textbf{6.00} &    \textbf{1.41} &  421.00 &   98.59 \\

    \midrule
          \multicolumn{1}{l}{} & \multicolumn{12}{c}{Vertical acceleration $a_z$} \\
          
           \multicolumn{1}{l}{}& \multicolumn{4}{c}{$Ts{\rm 1}$}       & \multicolumn{4}{c}{$Ts{\rm 2}$}       & \multicolumn{4}{c}{$Ts{\rm 3}$} \\
         \multicolumn{1}{l}{} & \multicolumn{1}{l}{$C$ [-]} & \multicolumn{1}{l}{$R_{\mathrm{c}}$ [\%]} & \multicolumn{1}{l}{$N$ [-]} & \multicolumn{1}{l}{$R_{\mathrm{nc}}$ [\%]} & \multicolumn{1}{l}{$C$ [-]} & \multicolumn{1}{l}{$R_{\mathrm{c}}$ [\%]} & \multicolumn{1}{l}{$N$ [-]} & \multicolumn{1}{l}{$R_{\mathrm{nc}}$ [\%]} & \multicolumn{1}{l}{$C$ [-]} & \multicolumn{1}{l}{$R_{\mathrm{c}}$ [\%]} & \multicolumn{1}{l}{$N$ [-]} & \multicolumn{1}{l}{$R_{\mathrm{nc}}$ [\%]} \\
         \midrule
    $\pm a_{z,\mathrm{ND}}$ & \textbf{194.00} &  \textbf{32.07} & 411.00 &  67.93 &  51.00  & 15.99 & 268.00 &  84.01  & 19.00  &  4.45 & 408.00 & 95.55 \\
    $\pm a_{z,\mathrm{AG}}$ & \textbf{38.00} &  \textbf{6.28} & 567.00 &   93.72  &  6.00  &  1.88 &  313.00   & 98.12    &4.00    &0.94  &423.00   &99.06 \\
    
    \bottomrule
    \end{tabular}%
    \end{adjustbox}
  \label{tab:thresholding_table}%
\end{table}%
An inspection of acceleration signals $a_y$ for all test sections shows that at all test sections signals exceed the public transportation threshold significantly. The number of detected critical sections are 221, 163 and 50 corresponding to $36.53\%$, $51.10\%$, $11.71\%$, for $Ts{\rm 1}$,  $Ts{\rm 2}$, and  $Ts{\rm 3}$, respectively. For normal driving, numbers are significantly lower: $2.81\%$ on  $Ts{\rm 1}$, $15.00\%$ on $Ts{\rm 2}$ and $1.41\%$ on $Ts{\rm 3}$ are classified as critical. For aggressive driving, only $Ts{\rm 2}$ and $Ts{\rm 3}$ have shown 1 and 6 critical sections, respectively, corresponding to $1.00\%$ and $1.41\%$. Finally, the results of $a_z$ show several signal peaks exceeding the thresholds throughout all three test sections. Especially $Ts{\rm 1}$ shows 194 critical sections for $\pm a_{z,\mathrm{ND}}$ and 38 critical sections considering the AG threshold. The relative numbers correspond to $32.07\%$ and $6.28\%$, respectively. $Ts{\rm 2}$ and $Ts{\rm 3}$ show 51 ($15.99\%$) and 19 ($4.45\%$) critical sections for normal driving, respectively. For aggressive driving, $Ts{\rm 2}$ and $Ts{\rm 3}$ show 6 and 4 sections, respectively, corresponding to 1.88\% and 0.94\%. 

As the thresholding method does not combine the acceleration signals from all three directions, it is not possible to compare the comfort of test sections with one final performance metric. Nevertheless, the results show that $Ts{\rm 3}$ contains significantly less critical sections for all three acceleration signals $a_x$, $a_y$, and $a_z$. $Ts{\rm 1}$ shows the most criticalities for $a_x$ and $a_z$, whereas $Ts{\rm 2}$ presents the most critical sections for $a_y$. 

\begin{figure}[!b]
    \centering
    \includegraphics[width=0.6\textwidth]{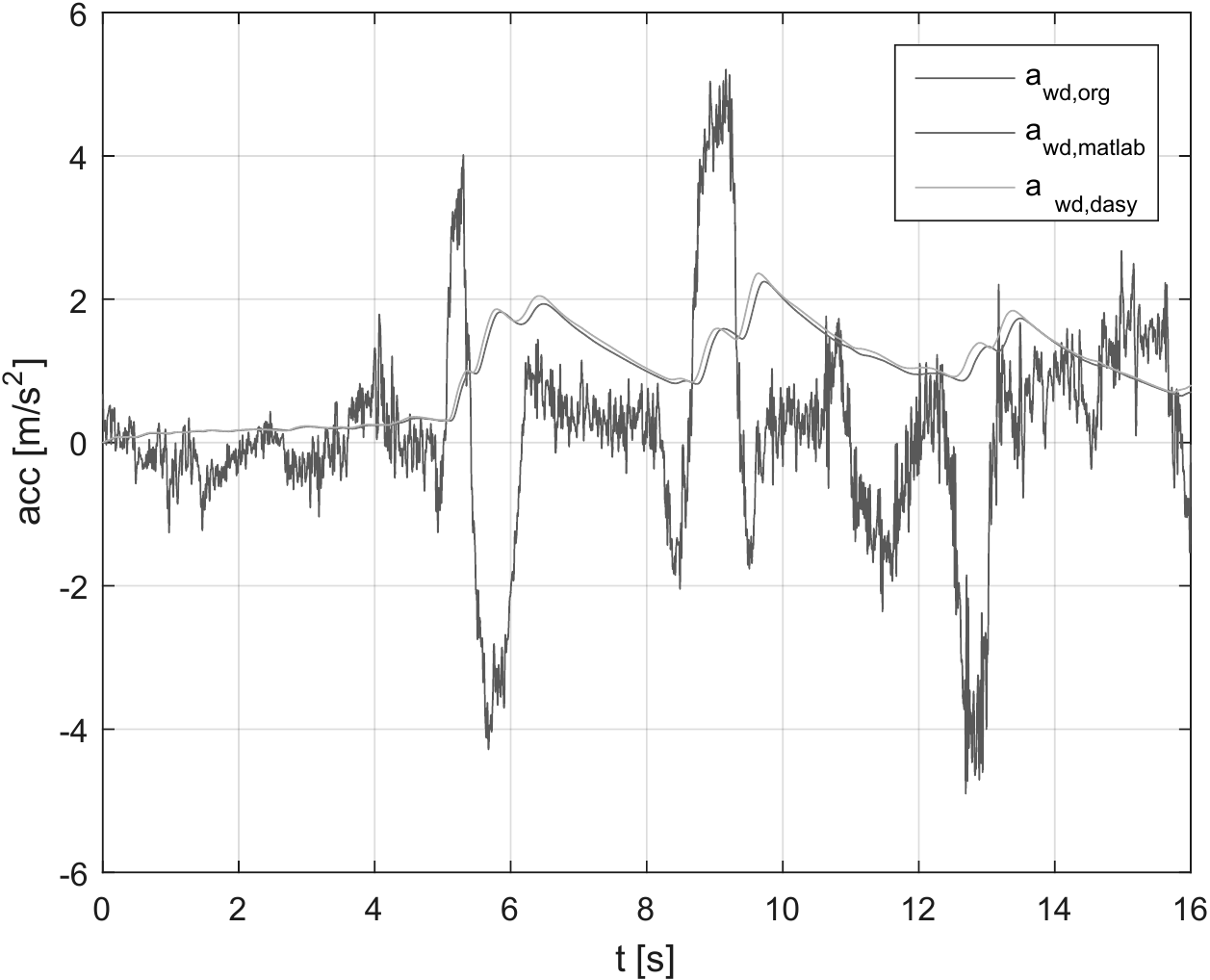}
    \caption{Validation of the ISO 2631 MATLAB implementation by comparing the frequency weighting with an add-on from DASYLab.}
    \label{fig:validation_iso}
\end{figure}

Secondly, we apply the ISO 2631 method to the derived simulation results. First, the implementation of the ISO 2631 for comfort evaluation needs to be validated. Hence, a frequency weighting add-on from software DASYLab has been utilized. A sample acceleration signal with time length of $t = 16$sec was utilized. Let $a_{\rm we,org}$ be the original acceleration signal, $a_{\rm we,matlab}$ the frequency weighted signal of MATLAB implementation, and $a_{\rm we,dasy}$ the output signal from DASYLab. The signals and frequency weighting results are depicted in Fig.~\ref{fig:validation_iso}. The output signals show that the implementation is valid and can be utilized for automated frequency weighting, which is a manual procedure in other software (e.g.,\ DASYLab). Nevertheless, it should be pointed out that a small time drift can be observed in Fig.~\ref{fig:validation_iso}, which is due to the different integration methods used. Because of computational limitations, MATLAB numerical integration was implemented as a convolution integral, whereas DASYLab uses a different approach.

\begin{figure}[t]
	\centering
	\includegraphics[width=0.65\textwidth]{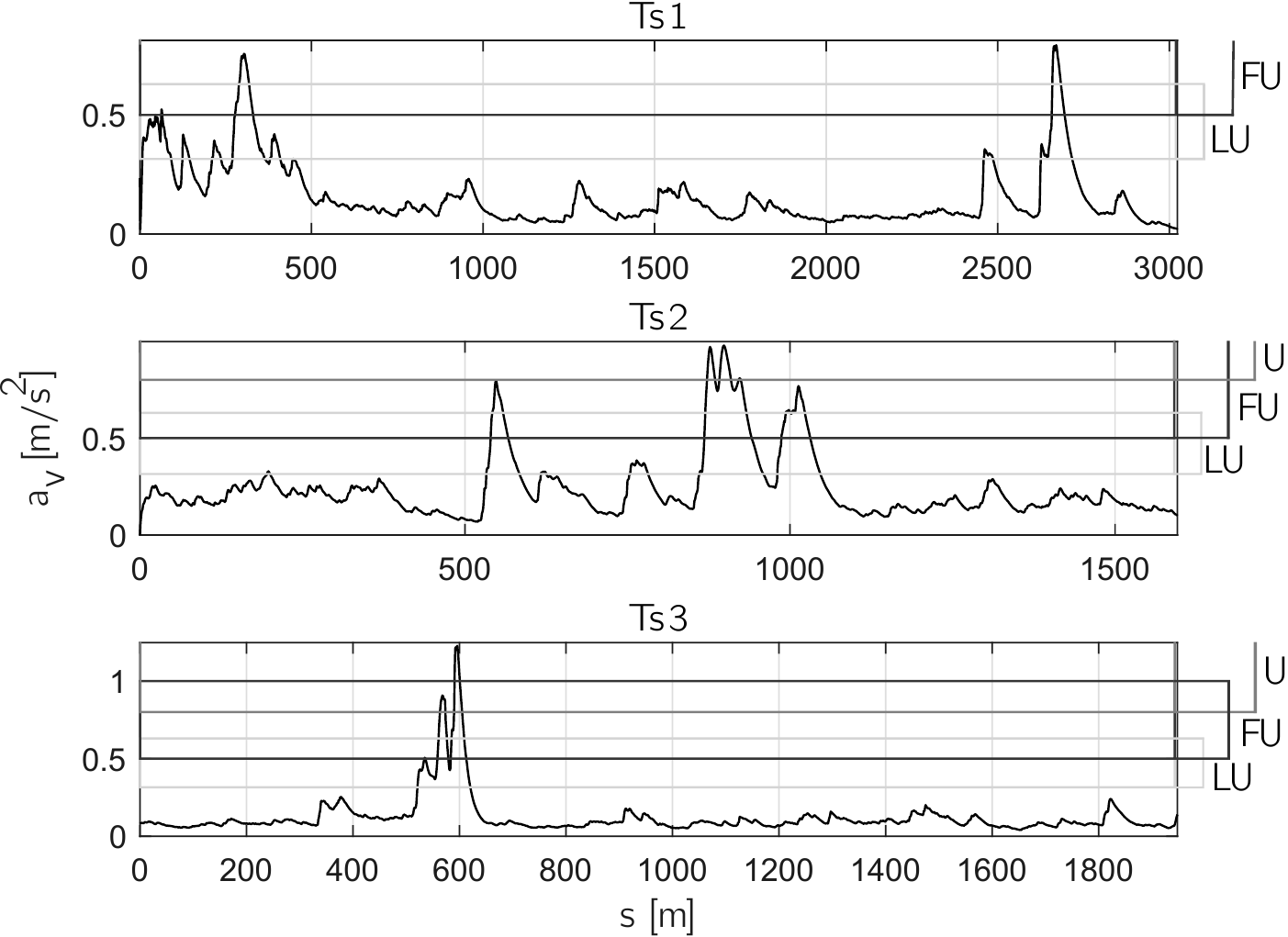}
	\caption{Ride comfort estimate with the ISO 2631 for the test sites $\mathrm{Ts}_{\rm 1}$, $\mathrm{Ts}_{\rm 2}$, and $\mathrm{Ts}_{\rm 3}$.}
	\label{fig:ride_comfors}
\end{figure}

In the following, the automated post-processing has been again applied to the 12.000 simulation outputs by filtering, integrating, and combining the signals according to the procedure described in Section~\ref{sec:iso_comfort}. Moreover, the time-dependent signals are transferred to the space domain in order to identify the critical sections. The final frequency-weighted acceleration signals are shown in Fig.~\ref{fig:ride_comfors} together with the recommended thresholds from ISO 2631. It can be seen that several signal peaks exceed the corresponding thresholds, thus indicating a negative ride comfort. Furthermore, the signal magnitude is an indicator of test site's smoothness; e.g., results of $Ts{\rm 2}$ in comparison to $Ts{\rm 3}$ show that the average signal magnitude below the threshold level LU is lower on $Ts{\rm 3}$. Note that the threshold for the probability of vibration perception ($0.01$m/s$^2$ to $0.02$m/s$^2$) is not shown in the results, as the magnitude is higher throughout the investigations, meaning that the perception of vibrations will always be present for the occupant.

To quantify the signal peaks, critical sections are determined per test site and shown in Table~\ref{tab:rid_comf_res}. Every site is processed sequentially for critical sections, again with a search window defined by $l_{\rm cr} = 5$m. It can be shown that $Ts{\rm 1}$ has the highest number of little uncomfortable (LU) sections (54 sections/8.92\% of the test site), followed by $Ts{\rm 2}$ and  $Ts{\rm 3}$ with 20 and 8 sections, respectively. $Ts{\rm 2}$ has the highest number of fairly uncomfortable (FU) and uncomfortable (U) sections, with 21 (6.58\% of the test site) and 8 (2.51\% of the test site), respectively. Consequently, the highest ratio of non-comfortable sections is detected on $Ts{\rm 2}$, followed by $Ts{\rm 1}$ and $Ts{\rm 3}$. Note that no sections were classified as very (VU) or extremely uncomfortable (EU).

\begin{table}[!b]
  \centering
  
  	\caption{Detected critical comfort road sections based on the ISO 2631 method for all test sites with a critical length $l_{\rm cr} = 5$m. The categories are defined as follows: LU=little uncomfortable, FU = fairly uncomfortable, VU=very uncomfortable and EU=extremely uncomfortable.}
	\begin{adjustbox}{width=\columnwidth,center}
    \begin{tabular}{lrrrrrrrrrrrr}
    \toprule
          & \multicolumn{12}{c}{Frequency-weighted acceleration $a_v$} \\
          
          & \multicolumn{4}{c}{$Ts{\rm 1}$}       & \multicolumn{4}{c}{$Ts{\rm 2}$}       & \multicolumn{4}{c}{$Ts{\rm 3}$} \\
          
          & \multicolumn{1}{l}{$C$ [-]} & \multicolumn{1}{l}{$R_{\mathrm{c}}$ [\%]} & \multicolumn{1}{l}{$N$ [-]} & \multicolumn{1}{l}{$R_{\mathrm{n}}$ [\%]} & \multicolumn{1}{l}{$C$ [-]} & \multicolumn{1}{l}{$R_{\mathrm{c}}$ [\%]} & \multicolumn{1}{l}{$N$ [-]} & \multicolumn{1}{l}{$R_{\mathrm{n}}$ [\%]} & \multicolumn{1}{l}{$C$ [-]} & \multicolumn{1}{l}{$R_{\mathrm{c}}$ [\%]} & \multicolumn{1}{l}{$N$ [-]} & \multicolumn{1}{l}{$R_{\mathrm{n}}$ [\%]} \\
          \midrule
            LU & \textbf{54.00}  &  \textbf{8.92} &  551.00 &  91.08    &     20.00    &     6.27 & 319.00 & 93.73   &  	 8.00  &  	  1.87 & 420.00 &  98.13 \\
            FU & 19.00  &  3.14 &  586.00 &  96.86    &     \textbf{21.00}    &     \textbf{6.58} & 319.00 & 93.42   &     6.00  &      1.40 & 427.00 &  98.60 \\
            U &   0.00  &  0.00 &  605.00 & 100.00    &     \textbf{8.00}     &     \textbf{2.51} & 311.00 & 97.49  &     5.00  &      1.17 & 427.00 &  98.83 \\
            VU &  0.00  &  0.00 &  605.00 & 100.00    &     0.00     &     0.00 & 319.00 & 100.00  &     0.00  &      0.00 & 427.00 & 100.00 \\
			EU &  0.00  &  0.00 &  605.00 & 100.00    &     0.00     &     0.00 & 319.00 & 100.00  &     0.00  &      0.00 & 427.00 & 100.00 \\
    \bottomrule
    \end{tabular}%
    \end{adjustbox}
  \label{tab:rid_comf_res}%
\end{table}%

\begin{figure}[!t]
    \centering
  \subfloat[]{%
       \includegraphics[width=0.51\linewidth]{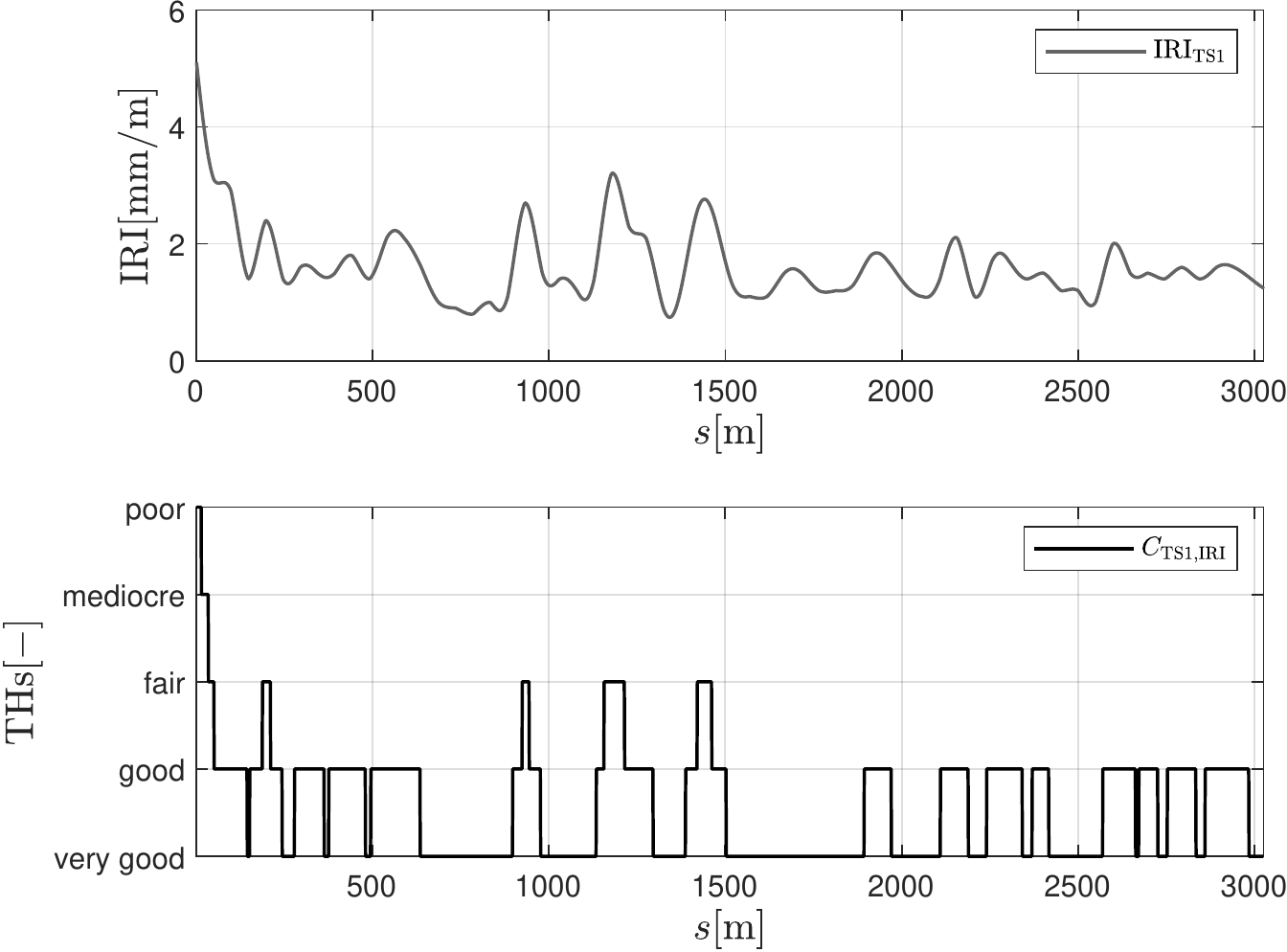}}
    \\
  \subfloat[]{%
       \includegraphics[width=0.51\linewidth]{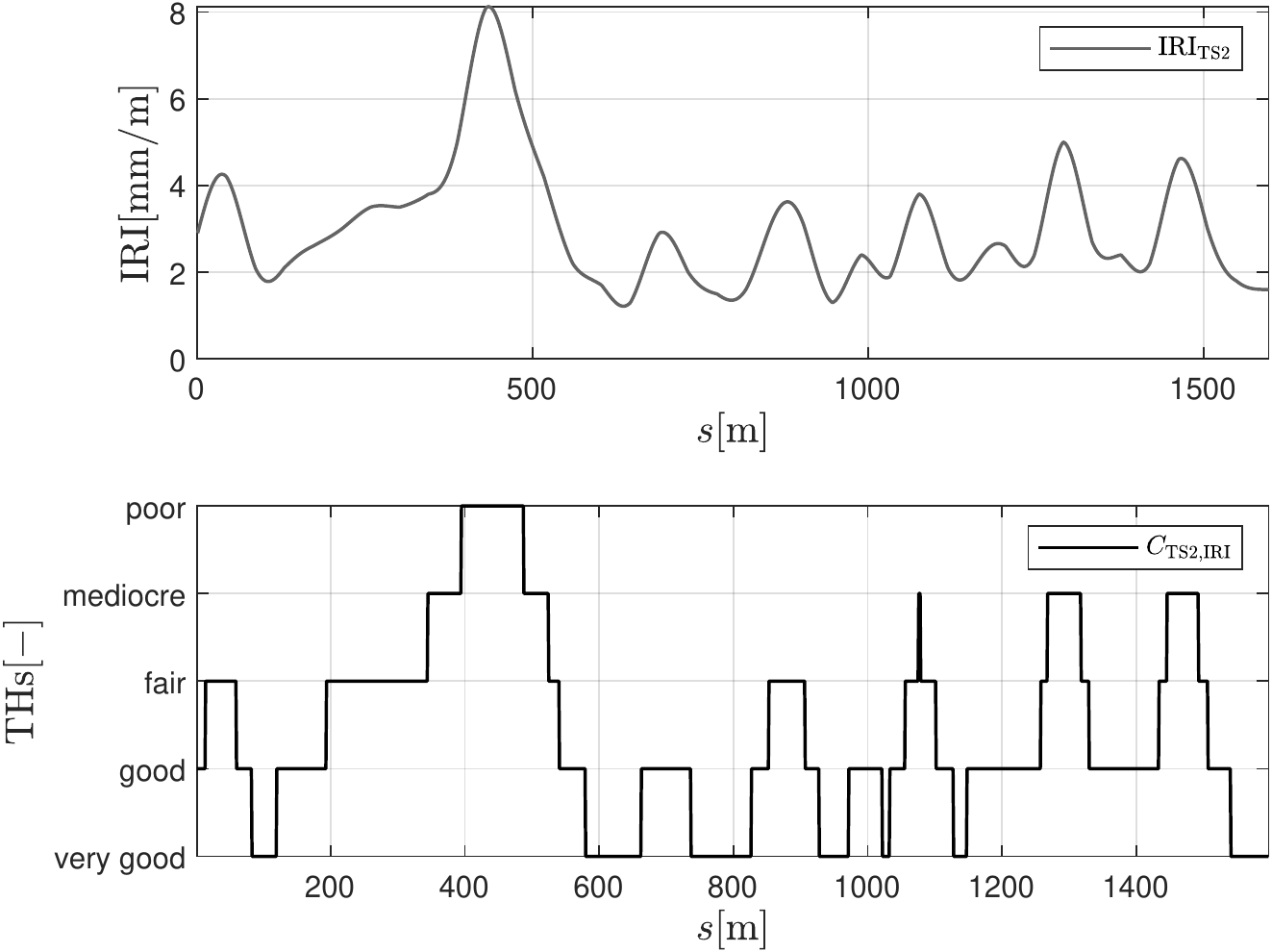}}
    \\
  \subfloat[]{%
        \includegraphics[width=0.51\linewidth]{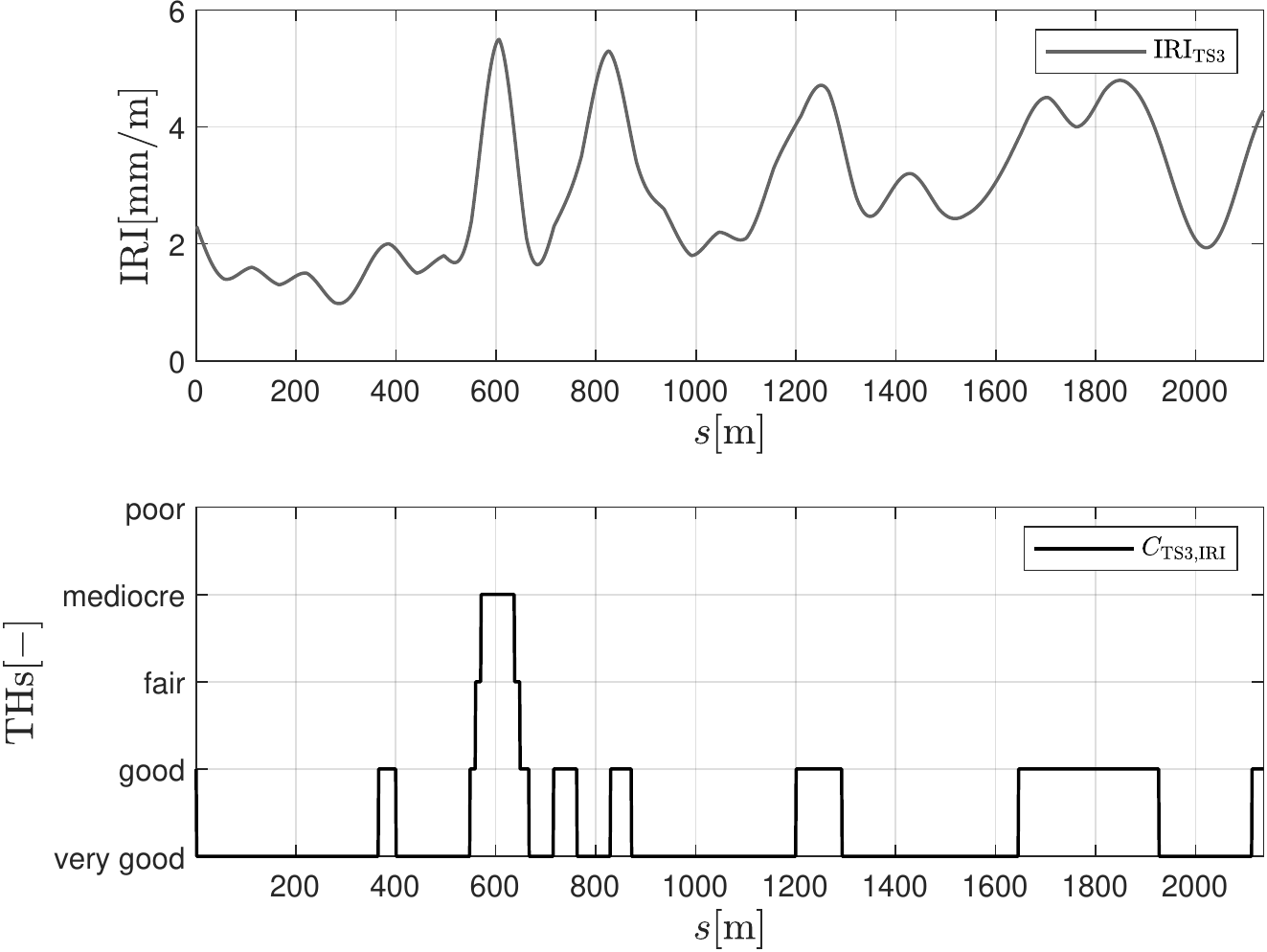}}
    \\
  \caption{IRI comfort classification results for the accelerations of $Ts{\rm 1}$, $Ts{\rm 2}$, and $Ts{\rm 3}$.}
  \label{fig:IRI_thresholds} 
\end{figure}

Finally, we utilize the empirical measurements, which allow for the derivation of IRI signals. Data were measured with a test vehicle from the Austrian Institute of Technology (AIT). The vehicle is equipped with a system (as introduced theoretically in Section~\ref{sec:IRI_concept}) that allows a high-precision derivation of IRI. Note that IRI measurements were taken right before the vehicle dynamics measurements (utilized for model parameters optimization in this work). Hence, it can be guaranteed that the same road conditions are present on $Ts{\rm 1}$, $Ts{\rm 2}$, and $Ts{\rm 3}$. The measurement equipment of the test vehicle from AIT allows for an IRI resolution of 50 meters. Therefore, we interpolate the data between two measurement samples. To derive ride comfort results, the speed-dependent thresholds from the work of~\cite{ref:Yu_IRI_comfort_THs} are considered. To apply the semi-dynamic thresholding, and as the correct threshold for an IRI sample is a function of speed, all simulation runs' average speed profiles are utilized. Figure~\ref{fig:IRI_thresholds} depicts the results for $Ts{\rm 1}$, $Ts{\rm 2}$, and $Ts{\rm 3}$. 

As depicted in Fig.~\ref{fig:IRI_thresholds}(a) $Ts{\rm 1}$ overall shows very good and good ride quality. Only one signal peak at the beginning of the test section shows poor quality, and four signal peaks are classified with the category fair. $Ts{\rm 3}$ shows a similar result with overall very good quality. Only a few signal peaks show good quality, and one signal peak identifies mediocre ride quality. Contrary to $Ts{\rm 1}$ and $Ts{\rm 3}$, $Ts{\rm 2}$ shows several signal peaks identifying fair, mediocre, and even poor ride quality. Also, Fig.~\ref{fig:IRI_thresholds}(b) shows long signal peaks in space, meaning that the number of critical sections is significantly higher than on the other test sections. To quantify the visual inspection, Table~\ref{tab:rid_comf_res2} presents the classification results for critical sections again with a length of $l_{\rm cr} = 5$m. Note that the category very good (VG) is not shown as all signals not classified by category good (G) to poor (P) are obviously a member of VG. 

\begin{table}[!b]
  \centering
  
  	\caption{Detected critical comfort road sections based on the IRI method for all test sites with a critical length $l_{\rm cr} = 5$m. The categories are defined as follows: G=good, F=fair, M=mediocre, P=poor.}
	\begin{adjustbox}{width=\columnwidth,center}
    \begin{tabular}{lrrrrrrrrrrrr}
    \toprule
          & \multicolumn{12}{c}{IRI signals $\mathrm{IRI}$} \\
          
          & \multicolumn{4}{c}{$Ts{\rm 1}$}       & \multicolumn{4}{c}{$Ts{\rm 2}$}       & \multicolumn{4}{c}{$Ts{\rm 3}$} \\
          
          & \multicolumn{1}{l}{$C$ [-]} & \multicolumn{1}{l}{$R_{\mathrm{c}}$ [\%]} & \multicolumn{1}{l}{$N$ [-]} & \multicolumn{1}{l}{$R_{\mathrm{n}}$ [\%]} & \multicolumn{1}{l}{$C$ [-]} & \multicolumn{1}{l}{$R_{\mathrm{c}}$ [\%]} & \multicolumn{1}{l}{$N$ [-]} & \multicolumn{1}{l}{$R_{\mathrm{n}}$ [\%]} & \multicolumn{1}{l}{$C$ [-]} & \multicolumn{1}{l}{$R_{\mathrm{c}}$ [\%]} & \multicolumn{1}{l}{$N$ [-]} & \multicolumn{1}{l}{$R_{\mathrm{n}}$ [\%]} \\
          \midrule
			G &   30.00    &	 4.96  	&	575.00    & 	  95.04   &	   \textbf{68.00}  &      \textbf{21.32}  &		251.00   &	  78.68   &      4.00   &      0.94  	&	423.00    &    99.06 \\
			F &    4.00    &	 0.66  	&	601.00    &    99.34      &  \textbf{36.00}   &		\textbf{11.29}  	&	283.00   	&  88.71  & 	   13.00     &    3.04  	&	414.00   	&  96.96 \\
			M &    3.00    &	 0.50  	&	602.00    &	  99.50   	  & \textbf{18.00}    &	 \textbf{5.64}  	&	301.00   &	  94.36      &   0.00    &     0.00  	&	427.00  	& 100.00 \\
			P &	0.00    &    0.00   &     0.00    &     0.00      &   0.00   &      0.00    &     0.00    &     0.00   &      0.00   &      0.00      &   0.00      &   0.00 \\
	\bottomrule
    \end{tabular}%
    \end{adjustbox}
  \label{tab:rid_comf_res2}%
\end{table}%

The category G identifies 30, 68 and 4 sections on $Ts{\rm 1}$, $Ts{\rm 2}$, and $Ts{\rm 3}$, respectively; corresponding to $4.96\%$, $21.32\%$, and $0.94\%$. For categories F, M and P, $Ts{\rm 1}$ shows low critical classification ratios, namely, $0.66\%$, $0.50\%$ and $0.00\%$. Also, $Ts{\rm 2}$ only classifies $3.04\%$ as F and $0.00\%$ as M and P. As already identified by ISO 2631 method, $Ts{\rm 2}$ shows the highest number of critical comfort road sections: $11.29\%$ are identified as F and $5.64\%$ as M; no sections are classified as P. 

A final comparison of the proposed methods shows that the thresholding method draws a different picture than the ISO 2631 and the IRI methods. The benchmark, i.e., IRI method, shows for all categories the highest number of criticalities on $Ts{\rm 2}$ (68, 36, 18 and 0 critical sections for categories G, F, M, and P). Contrary, the thresholding shows for the acceleration $a_x$ the most criticalities on  $Ts{\rm 1}$ with 24 and 21 detected sections for the PT and ND threshold, respectively. Thresholding $a_y$ shows for PT and ND $Ts{\rm 2}$ as the most critical, although the result for AG is slightly higher (6 critical sections, 1.41\%) on  $Ts{\rm 3}$. For $a_z$, the most critical sections are detected on $Ts{\rm 1}$ with a number of 194 or 32.07\% of the whole test section. 

The ISO 2631 method results show the most detected sections exceeding the threshold for LU for $Ts{\rm 1}$. Apart from LU, the number of most detected critical sections for FU and VU occur on $Ts{\rm 2}$ with 21 and 8 or 6.58\% and 2.51\%. The algorithm detects no EU sections. Consequently, the detections of the ISO 2631 and IRI method are substantially similar, although the magnitude of detections is higher with the IRI method. 

Although the thresholding procedure gives different results and draws a different picture for the accelerations $a_x$, $a_y$, and $a_z$, the method is useful for the detection and/or mitigation of uncomfortable longitudinal or latitudinal vehicle motions. Uncomfortable longitudinal movements are detectable by $a_x$ exceeding a chosen threshold for uncomfortable acceleration/braking. By $a_y$ exceeding a chosen threshold, uncomfortable turns can be detected. Besides, $a_z$ gives insights into the roughness of the road surface. 


\section{CONCLUSION}
\label{sec:diss_concl}
This paper presents a novel method for deriving ride comfort data to improve AVs' route and motion planning systems. The proposed methodology provides a general framework for comfort evaluation and compares different methodologies of comfort derivation. The high-resolution outputs can be utilized for additional information to data fusion systems, stimulating AVs' efficient and passenger-oriented development. In addition, the modular structure of the framework allows for its integration to software or hardware in the loop tests that provide vehicle dynamics data as an output. The presented case study demonstrates that the selected test sites indeed show acceleration signal peaks that are critical concerning comfort. All three presented methodologies, i.e., thresholding, ISO 2631, and IRI method, identify critical comfort road sections. Although ISO 2631 and IRI results are substantially similar, the thresholding method can be utilized meaningfully to detect critical vehicle motions. The work suggests to utilize methodologies such as ISO 2631 as a minimum standard for AVs to quantify and mitigate negative ride comfort; i.e., the standalone thresholding of acceleration data can be misleading and over- or underestimate situations that potentially lead to negative ride comfort. The proposed framework can be implemented in future data fusion systems or added to a digitized map utilized by AVs, allowing for more efficient planning of routes and motions and avoidance of negative ride comfort experiences. However, a limitation of the proposed work is the necessity of accurate road data availability required to model the road surface in the simulation environment.

Future research should investigate the application of the derived comfort results to an AV's driver model to assess the different  comfort evaluation methods' performance in real-time applications. Moreover, the parameters optimization results could be further improved by comparing different available solvers for non-linear problems. Also, several methods for defining the optimal initial values or application of a random restart under certain conditions could be considered. Finally, another interesting future research direction would be the application the obtained ride comfort data to data fusion systems of AVs and assessment the performance.



\bibliographystyle{elsarticle-harv}
\bibliography{bibliography}

\newpage

\end{document}